%% file: main.tex
\documentclass[10pt,twocolumn,letterpaper]{article}

\usepackage[pagenumbers]{cvpr} % To force page numbers, e.g. for an arXiv version

\input{preamble}

\definecolor{cvprblue}{rgb}{0.21,0.49,0.74}
\usepackage[pagebackref,breaklinks,colorlinks,allcolors=cvprblue]{hyperref}
\def\confName{CVPR}
\def\confYear{2026}
\title{Diagnose, Correct, and Learn from Manipulation Failures via Visual Symbols}
\author{
    Xianchao Zeng$^{1,2*}$ \quad Xinyu Zhou$^{2,3*}$ \quad Youcheng Li$^{1,2}$ \quad Jiayou Shi$^4$ \quad Tianle Li$^4$  \\ Liangming Chen$^{3 \dagger}$ \qquad Lei Ren$^{1 \dagger}$ \qquad Yong-Lu Li$^{2,4 \dagger}$ \\
    $^1$Beihang University \quad $^2$Shanghai Innovation Institute \quad 
    \\ $^3$Southern University of Science and Technology \quad $^4$Shanghai Jiao Tong University
    \\
    {\tt\small xianchao\_zeng@buaa.edu.cn \qquad zhouxy2023@mail.sustech.edu.cn \qquad yonglu\_li@sjtu.edu.cn}
}
\usepackage[accsupp]{axessibility}  % Improves PDF readability for those with disabilities.
\usepackage[most]{tcolorbox}
\usepackage{setspace} % optional if you want tighter or looser spacing

\usepackage{graphicx}
\usepackage{caption}
\usepackage{enumitem}

\newtcolorbox{llmprompt}[2][]{
    enhanced,
    sharp corners,
    colback=#2!8,        % main background (light)
    colframe=black,      % border color
    coltitle=white,      
    fonttitle=\bfseries,
    title=#1,
    attach boxed title to top left={
        xshift=0mm,
        yshift*=-1mm
    },
    boxed title style={
        colback=#2!40,   % title bar background (dark)
        arc=0mm,
        top=2pt,
        bottom=2pt,
        left=3pt,
        right=3pt
    },
    boxrule=0.75pt,
    left=4pt,
    right=4pt,
    top=6pt,
    bottom=6pt,
}

\begin{document}
\maketitle

\let\thefootnote\relax\footnotetext{
  \begin{tabular}{@{}l@{ }l}
    * & Equal contribution.\\
    $\dagger$ & Corresponding authors.
  \end{tabular}
}

\input{sec/0_abstract}    
\input{sec/1_intro}
\input{sec/2_related_work}
\input{sec/3_design}
\input{sec/4_experiment}
\input{sec/5_conclusion}

{
    \small
    \bibliographystyle{ieeenat_fullname}
    \bibliography{main}
}

\input{sec/X_suppl}

\end{document}

%% file: preamble.tex
\usepackage{microtype}

\usepackage[table]{xcolor}

%% file: sec/0_abstract.tex
\begin{abstract}
Vision-Language-Action (VLA) models have recently achieved remarkable progress in robotic manipulation, yet they remain limited in failure diagnosis and learning from failures. Additionally, existing failure datasets are mostly generated programmatically in simulation, which limits their generalization to the real world. In light of these, we introduce \textbf{ViFailback}, a framework designed to diagnose robotic manipulation failures and provide both textual and visual correction guidance. 
Our framework utilizes explicit visual symbols to enhance annotation efficiency. We further release the ViFailback dataset, a large-scale collection of 58,128 Visual Question Answering (VQA) pairs along with their corresponding 5,202 real-world manipulation trajectories. Based on the dataset, we establish ViFailback-Bench, a benchmark of 11 fine-grained VQA tasks designed to assess the failure diagnosis and correction abilities of Vision-Language Models (VLMs), featuring ViFailback-Bench Lite for closed-ended and ViFailback-Bench Hard for open-ended evaluation. 
To demonstrate the effectiveness of our framework, we built the ViFailback-8B VLM, which not only achieves a significant overall performance improvement on ViFailback-Bench but also generates visual symbols for corrective action guidance. 
Finally, by integrating ViFailback-8B with a VLA model, we conduct real-world robotic experiments demonstrating its ability to assist the VLA model in recovering from failures. \href{https://x1nyuzhou.github.io/vifailback.github.io/}{Project Website.}
\end{abstract}

%% file: sec/1_intro.tex
\section{Introduction}
\label{sec:intro}

\begin{figure}[t]
    \centering
    \includegraphics[width=0.9\columnwidth]{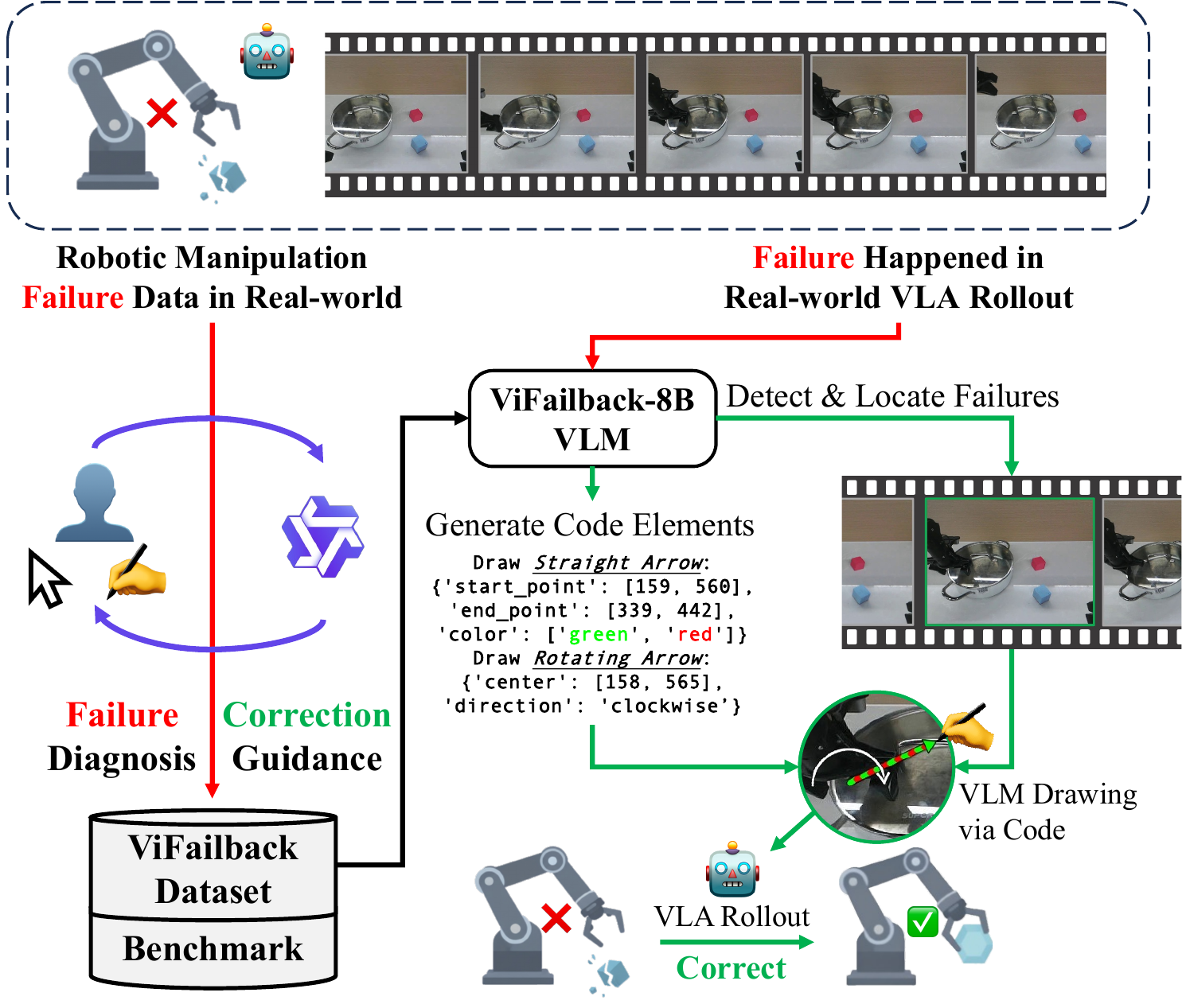}
    \caption{The pipeline leverages real-world failure data to build a dataset and train ViFailback-8B for failure diagnosis and correction.}
    \label{fig:first}
\end{figure}

Imitation learning has emerged as a simple but efficient paradigm for robot control, allowing robots to directly master diverse manipulation skills from expert demonstration data. Recently, as large-scale, high-quality robotic datasets have become available, Vision-Language-Action (VLA) models \cite{black2025pi_, kim2024openvla, liu2024rdt} have shown impressive performance in robotic manipulation. 
However, when deployed in the real world, these models inevitably encounter out-of-distribution (OOD) conditions, where input observations deviate from the training data distribution. In such cases, the generated actions can lead to failures. Therefore, enabling robotic systems to diagnose, learn, and recover from failures is crucial for achieving robust general-purpose manipulation in open-world environments.

The latest progress in Vision-Language Models (VLMs) \cite{team2025robobrain, azzolini2025cosmos, huang2024copa, qi2025sofar} has enabled exceptional performance in robotic task planning, spatial reasoning, and trajectory prediction. However, these models often struggle to accurately analyze and correct failures in robotic manipulation tasks. Some recent works \cite{duan2024aha, dai2025racer, lu2025robofac} have focused on automatically generating large-scale failure datasets, typically by injecting perturbations in simulation, which are subsequently used to finetune VLMs and improve their failure diagnosis and correction performance. This strategy is fundamentally constrained by the sim-to-real gap, limiting its efficiency in real-world settings. In fact, during teleoperation data collection or policy rollout, it is common for robots to produce a certain amount of failure data alongside successful demonstrations. However, annotating such data simply and efficiently remains a significant challenge. 

To leverage real-world failure data, we propose a framework for robotic manipulation failure diagnosis and correction. As illustrated in Figure \ref{fig:first}, given a robotic manipulation video, human annotators can perform efficient and low-cost annotation by drawing visual symbols with a mouse and leveraging a VLM to assist the labeling process. 
We collected 5,202 real-world trajectories via teleoperation and policy rollouts, covering 100 distinct tasks and spanning four major categories of manipulation failures. Using the ViFailback framework, we generated the ViFailback dataset, comprising 58,128 Visual Question Answering (VQA) pairs tailored for robotic failure diagnosis and correction. 
Given this dataset, we finetuned Qwen3-VL-8B \cite{Qwen3-VL} to create ViFailback-8B and proposed the ViFailback-Bench for evaluating VLMs in both closed-ended and open-ended settings. 
Experiments show ViFailback-8B significantly outperforms other open-source and proprietary models on our benchmark. 
Furthermore, we deploy ViFailback-8B as an external supervisor during robot task execution, allowing it to intervene in policy execution to recover from failures, which is an essential component for a policy to truly learn from its failures. Compared with baselines without ViFailback-8B, this integration leads to an average performance improvement of 22.2\%. We summarize our contributions as follows:
\begin{itemize}[leftmargin=2em, topsep=0pt, itemsep=0pt]
    \item We propose ViFailback, a scalable framework for efficiently annotating real-world robotic failure videos with both textual explanations and visual symbols.
    \item We release a dataset with 58,128 high-quality VQA pairs for failure diagnosis and correction, spanning 11 different question types. We then establish the ViFailback-Bench to comprehensively evaluate the failure diagnosis and correction abilities of VLMs.
    \item Our experiments on ViFailback-Bench demonstrate the effectiveness of ViFailback in improving the diagnosis and correction capabilities of general VLMs for robotic failures. Moreover, our real-world experiments confirm that this leads to a measurable improvement in the policy's capacity to recover from failures.
\end{itemize}

%% file: sec/2_related_work.tex
\section{Related Work}
\subsection{Robotic Manipulation with Imitation Learning}
Imitation learning allows robots to acquire manipulation skills from expert demonstrations, showing immense potential in robot learning. Among recent advancements, Diffusion Policy \cite{chi2025diffusion} and its variants \cite{ze20243d, wang2024equivariant}, which use diffusion models to denoise sampled trajectories, have achieved performance comparable to expert demonstrations. Furthermore, Vision-Language-Action (VLA) models \cite{kim2024openvla, black2025pi_, liu2024rdt, zitkovich2023rt} pre-trained on large-scale robotics datasets have demonstrated impressive capabilities in general-purpose manipulation. To scale high-quality data, recent works \cite{xu2025exumi, luo2025human} leverage sophisticated hardware and shared-control frameworks to bolster collection efficiency. However, the policies trained exclusively on these successful datasets, despite their rapid skill acquisition, are prone to failure when encountering OOD scenarios. This exposes a crucial blind spot: VLAs are unable to identify, let alone correct failures. Consequently, the development of methods for robotic failure diagnosis and correction becomes imperative. \cite{gu2025safe, wang2025vlatest}

\subsection{Failure Detection and Recovery}
Recent efforts tackle robotic failure detection and correction using external Visual Language Models (VLMs). YAY \cite{shi2024yell} refines corrective instructions via human-in-the-loop feedback but faces scalability challenges. Other works \cite{duan2024aha, dai2025racer, lu2025robofac, lin2025failsafe} synthesize failure data in simulation, yet their utility is constrained by the sim-to-real gap, while Fail2Progress \cite{huang2025fail2progress} uses real failures primarily to bootstrap simulation for skill-effect modeling.
Meanwhile, several benchmarks study VLM reasoning in robotics: Robo2VLM \cite{chenrobo2vlm} provides 684k in-the-wild QA pairs for spatial and goal-conditioned reasoning; ManipBench \cite{zhao2025manipbench} evaluates low-level manipulation reasoning. These benchmarks assess what to do and how to do it, but not what went wrong and why.
In contrast, we build ViFailback-Bench, a large-scale real-world VQA dataset and benchmark for fine-grained failure reasoning. Although prior methods \cite{lu2025robofac, duan2024aha, liu2023reflect} grant VLMs some diagnostic or corrective abilities, their feedback remains mostly textual, limiting robotic recovery due to the fragile instruction-following abilities of current VLA models \cite{glossop2025cast, liu2023interactive}.

\subsection{Visual Prompts for Robot Learning} 
Recent progress has explored using visual prompts to guide robot policies. Early work introduced trajectory-conditioned models \cite{zheng2024tracevla, gu2023rt, sundaresan2024rt, mehta2025l2d2}, which overlay 2D end-effector trajectories onto images but cannot revise these trajectories when execution deviates. Later approaches adopt semantic visual prompts for instruction, such as keypoint and grid-cell cues in MOKA \cite{liu2024moka}, images of objects or scenes in VIMA \cite{jiang2022vima}, and hand-drawn symbolic representations (\eg, arrows, circles) in RoVI \cite{li2025robotic} and CrayonRobo \cite{li2025crayonrobo}. However, these symbolic prompts focus on initial guidance rather than real-time correction. In contrast, our method enables a VLM to generate both textual and visual prompts for failure recovery and provides a structured mechanism that guides VLA models to adjust their actions based on our designed visual symbols.

%% file: sec/3_design.tex
\begin{figure*}[htbp]
    \centering
    \includegraphics[width=0.91\textwidth]{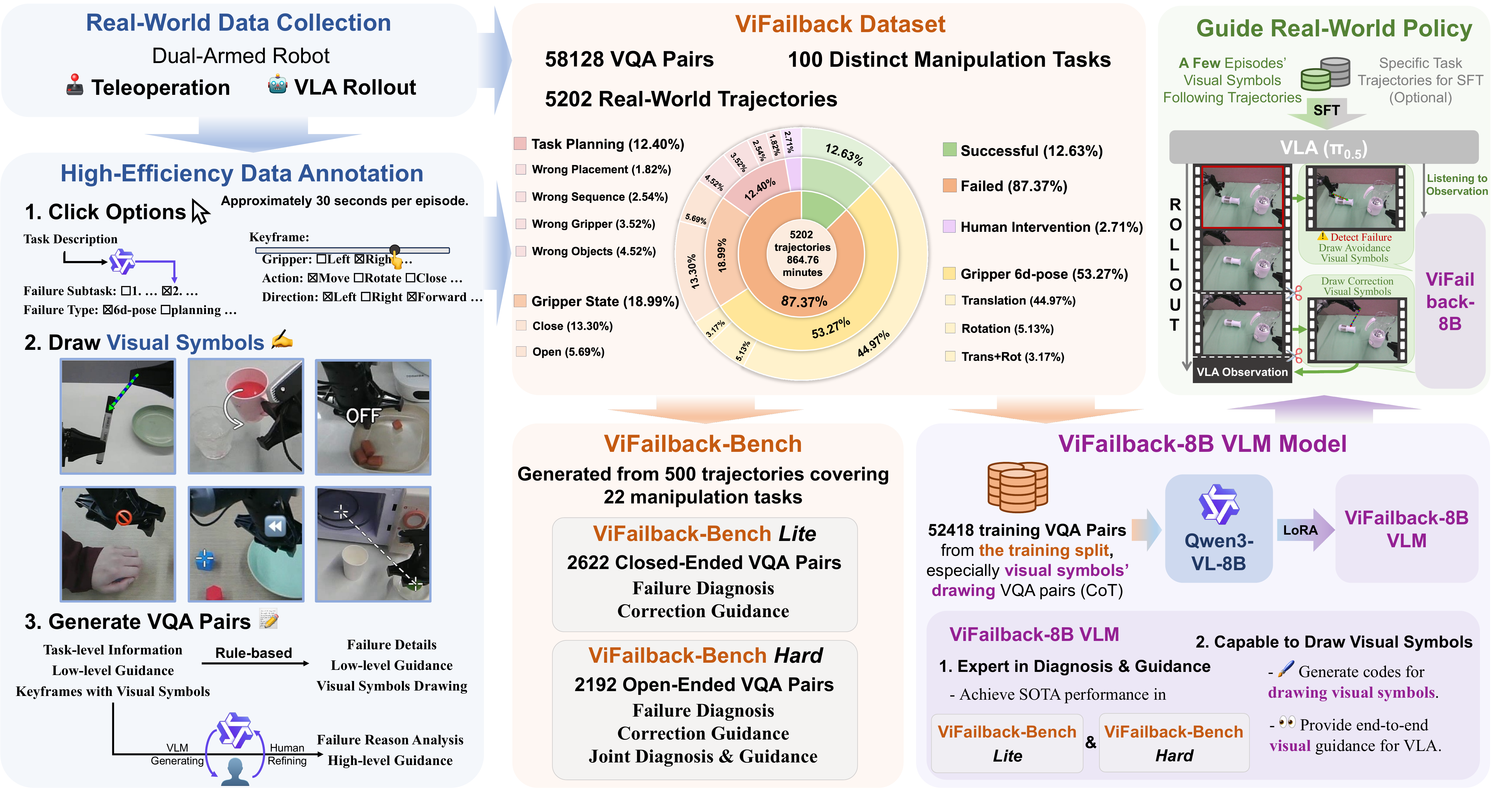}
    \caption{\textbf{Overview of ViFailback Framework.} \textbf{Left:} We collect real-world manipulation trajectories via teleoperation and policy rollout, then use our high-efficiency, visual-symbol-based annotation framework to generate VQA pairs for the dataset. \textbf{Middle:} Our dataset comprises 58,128 VQA pairs from 5,202 real-world trajectories. We extract ViFailback-Bench (Lite and Hard) from this dataset to evaluate VLM failure diagnosis and correction capabilities. \textbf{Right:} We finetune Qwen3-VL-8B on our VQA pairs to obtain ViFailback-8B. This model is deployed as an external supervisor to assist the policy in recovering from failures.}
    \label{fig:1}
\end{figure*}

\section{ViFailback}
We propose a framework named ViFailback (as illustrated in Figure \ref{fig:1}) for systematically annotating real-world robotic failure data. The framework decomposes the annotation process into two components: diagnosis and correction, and leverages visual symbols to facilitate semi-automated annotation. 
In this section, we detail the ViFailback annotation pipeline, including the design of visual symbols and task definition. Based on this framework, we construct the ViFailback dataset and establish the ViFailback-Bench, a benchmark that enables a comprehensive evaluation of VLMs in diagnosing and correcting robotic manipulation failures.

\subsection{Visual Symbols of ViFailback}
In contrast to failure data generated in simulation, real-world failure data requires extensive manual labeling. The annotation process is particularly inefficient for complex and abstract categories such as task planning failures, failure reason, and high-level corrective guidance, as it typically relies on manual textual descriptions. To address this, our framework employs visual symbols, which can be drawn directly onto video frames, to streamline the manual annotation process. Guided by these visual cues, VLM can then automate the generation of desired annotations for these complex categories, bypassing the need for manual textual labeling.

As shown in the left part of Figure \ref{fig:1}, we propose 7 distinct visual symbols to provide corrective action guidance. To clearly delineate their functions, we group these symbols into the following three categories: 

\textbf{Motion Symbols.} 
This category represents the translation and rotation of the robot end-effector.
\begin{itemize}[leftmargin=2.2em]
  \item \textbf{Colored Straight Arrow.} To enable 2D arrows to represent 3D spatial movement, we use color to indicate the direction of motion. The three colors used are \textcolor{red}{Red}, \textcolor{green}{Green}, and \textcolor{blue}{Blue}, corresponding directly to the three orthogonal directions: \textcolor{red}{Red} for forward-backward, \textcolor{green}{Green} for left-right, and \textcolor{blue}{Blue} for up-down.

  \item \textbf{Semi-circular Arrow.} The semi-circular arrow is used to indicate the rotation of the end-effector, while the pointing direction of the arrow indicates the expected rotation direction (clockwise or counterclockwise).
\end{itemize}

\textbf{Spatial Relation Symbols.} This category specifies the correct target object in the scene or the desired alignment between two objects.
\begin{itemize}[leftmargin=2.2em]
\item \textbf{Dual Crosshairs.} The dual crosshairs icon, linked by a dashed line, denotes that two targets are intended to be aligned.

\item \textbf{Crosshair.} The crosshair icon is used to highlight an ideal object or area in the video frame.
\end{itemize}

\textbf{State Symbols.} This category indicates the desired state of the target object.
\begin{itemize}[leftmargin=2.2em]
\item \textbf{ON/OFF Labels.} The ON/OFF labels are used to indicate the ideal state of the robot end-effector (open or closed). 

\item \textbf{Prohibition Icon.} The prohibition icon placed on the end-effector represents that the end-effector is expected to halt.

\item \textbf{Rewind Icon.} The rewind icon, placed on an end-effector or object, signifies that the specified component must return to a previous state.
\end{itemize}

In our annotation pipeline, we record the key elements for each visual symbol—such as its category, start points, and end points—so that the VLM can learn to draw visual correction guidance just as a human annotator would.

\begin{figure*}[htbp]
    \centering
    \includegraphics[width=0.93\textwidth]{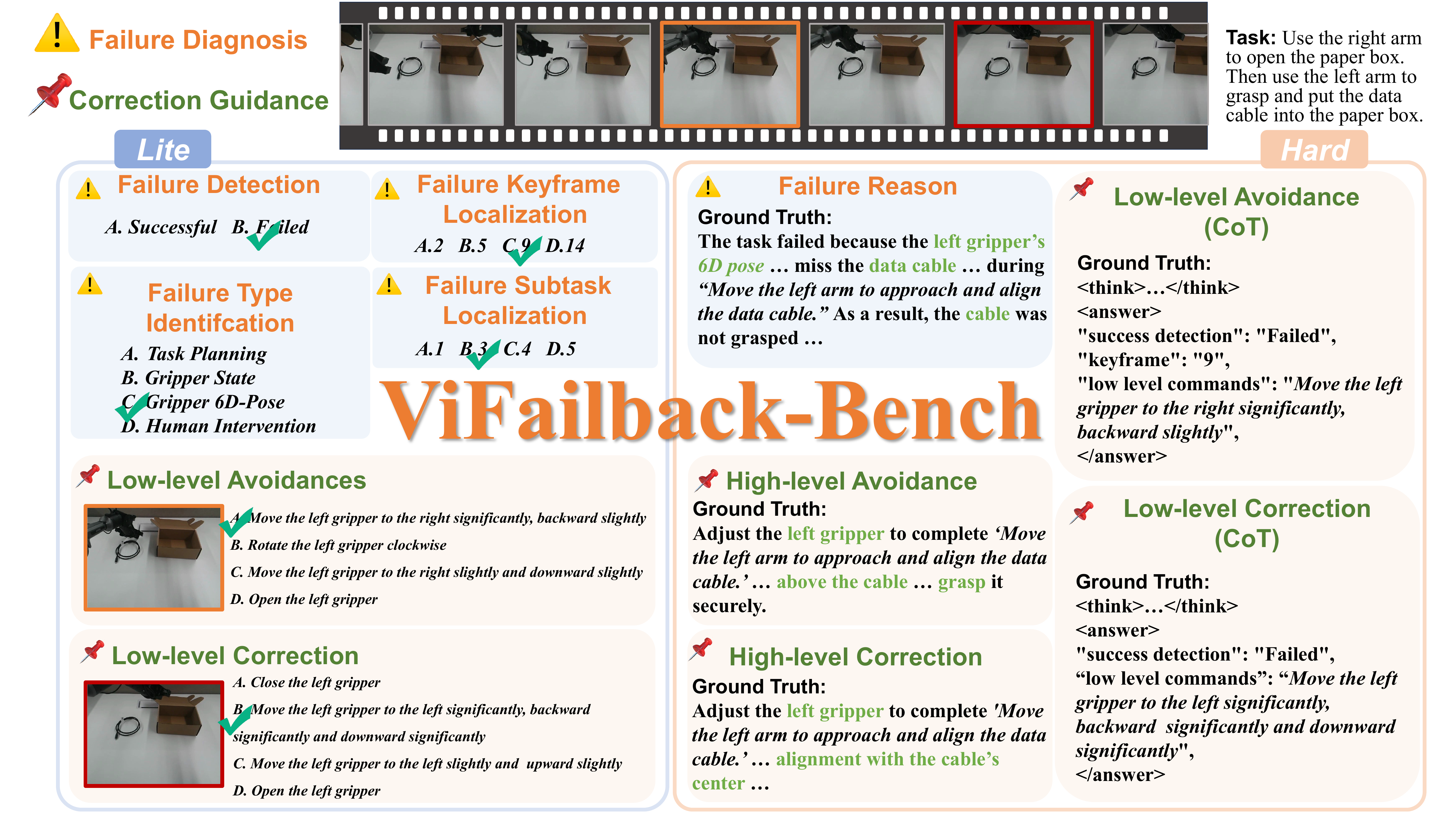}
    \caption{\textbf{An overview of ViFailback-Bench.} \textbf{Left:} The Lite benchmark uses closed-ended VQA to test VLM failure diagnosis (\eg, detection, localization) and low-level correction guidance. \textbf{Right:} The Hard benchmark uses open-ended VQA to test failure reason and high-level/CoT-based guidance.}
    \label{fig:2}
\end{figure*}

\subsection{Fine-Grained Task Definition}
\label{sec:task_definition}
In the ViFailback framework, failure analysis consists of two key components: failure diagnosis and corrective action guidance. The former involves identifying, localizing, and analyzing the cause of the failure; the latter provides the robot with the necessary actions or tasks to either avoid the impending failure or recover from it.

\textbf{Failure Diagnosis.} We define failure diagnosis as the comprehensive analysis of a robotic manipulation video, encompassing the following five components:
\begin{itemize}[leftmargin=2em, topsep=0pt, itemsep=0pt]
    \item \textit{Failure detection:} Determining whether the robotic manipulation task in the video was completed.
    \item \textit{Failure keyframe localization:} Localizing the keyframe signifying an impending failure.
    \item \textit{Failure subtask localization:} Localizing the subtask where the failure first began to occur.
    \item \textit{Failure type identification:} Identifying the type of failure that occurred. The failure types are categorized into 4 main types: (1) Task planning: Errors in the high-level task plan, such as targeting an incorrect interaction object or location, improper subtask sequencing, or the omission of a necessary subtask. (2) Gripper 6d-pose: The gripper fails to reach its correct position or orientation. (3) Gripper state: The gripper does not close or open properly, or its level of closure or opening is insufficient. (4) Human intervention: Disruptions from external forces that prevent task continuation. 
    \item \textit{Failure reason:} Reasoning about the failure's root cause and providing a detailed explanation.
\end{itemize}

\textbf{Corrective Action Guidance.} ViFailback requires models to provide textual and visual guidance for avoiding and correcting failures in robotic manipulation tasks.
\begin{itemize}[leftmargin=2em, topsep=0pt, itemsep=0pt]
    \item \textit{Low-level textual guidance:} Providing specific movement and rotation directions for the end-effector to avoid or correct the failure.
    \item \textit{High-level textual guidance:} Formulating high-level strategic advice (\eg, subtask plan reformulation) to recover from the failure.
    \item \textit{Visual guidance:} Providing visual guidance by overlaying visual symbols with integrated semantic information onto keyframes.
\end{itemize}

\subsection{Dataset and Benchmark}
We collected 5,202 robotic manipulation trajectories and their corresponding ego-centric videos from 100 diverse real-world tasks. Using the ViFailback framework, we thoroughly annotated detailed information for failure diagnosis and correction, culminating in the ViFailback dataset, which features 58,128 high-quality VQA pairs. Furthermore, to enable a comprehensive evaluation of VLMs in failure diagnosis and correction, we propose the ViFailback-Bench benchmark.

\subsubsection{Data Collection}
\label{sec:data_collection}
Our 100 designed tasks cover a diverse set of manipulation skills (\eg, place, pull, transfer, and pour). We collected trajectories using the ALOHA dual-arm robot platform \cite{fu2024mobile}, which includes 657 successful and 4,545 failed trajectories. Among these, 4,995 trajectories were gathered from human teleoperation, and the others were collected by executing the tasks with $\pi_{0.5}$ model \cite{black2025pi_}, a leading VLA model, which was finetuned on the successful teleoperated samples. The distribution of failure types across all collected trajectories is shown in the middle part of Figure \ref{fig:1}.

\subsubsection{Data Annotation Pipeline}
To ensure the high efficiency and quality of data annotation, we design a multi-stage data annotation pipeline as shown in Figure \ref{fig:1}. The pipeline consists of three main stages: 
(1) basic task semantic information filling, (2) textual guidance choosing and visual symbols drawing, and (3) open-ended description generation and refining.

In Stage 1, the annotators finish the failure diagnosis annotations (detection, keyframe localization, \etc.) through simple UI controls, like sliders and buttons with task descriptions decomposed into subtasks by Qwen2.5-Max \cite{bai2025qwen2}. 
In Stage 2, based on the selected keyframes, the annotators choose appropriate action correction options from predefined categories and draw visual symbols by dragging the mouse to indicate the corrective actions. 
In Stage 3, we prompt Qwen3-VL-235B \cite{Qwen3-VL} with all the annotated information and visual symbols to generate high-level descriptions for failure reasoning and high-level textual guidance. These outputs are then manually verified and refined to ensure high quality.

\subsubsection{ViFailback-Bench}
The ViFailback-Bench benchmark includes 500 trajectories across 22 distinct tasks. As depicted in Figure \ref{fig:2}, ViFailback-Bench comprises two complementary settings: ViFailback-Bench Lite and ViFailback-Bench Hard. 
The Lite version utilizes closed-ended VQA to assess core failure diagnosis abilities and low-level corrective action guidance grounded on provided keyframes. 
Then, the Hard version employs open-ended VQA to probe failure reasoning and correction capabilities. 
The low-level guidance tasks in the Hard setting are more complex, mandating the model to first detect and localize the failure, then output the guidance in a Chain-of-Thought (CoT) \cite{wei2022chain} format.

%% file: sec/4_experiment.tex
\section{Experiment}
In this section, we first conduct a comprehensive evaluation of 16 state-of-the-art models, encompassing both open-source and proprietary ones, on the ViFailback-Bench. To evaluate the effectiveness of our dataset, we then finetune Qwen3-VL-8B \cite{Qwen3-VL} on the ViFailback dataset to build ViFailback-8B, assessing its ability to enhance failure diagnosis and correction. Finally, we explore the real-world application of ViFailback-8B in assisting a VLA model with failure recovery to improve task success rates.

\subsection{Experiment Setup}

\textbf{Baseline.} We compare our model against 2 leading proprietary VLMs (GPT-4o \cite{hurst2024gpt} and Gemini-2.5-Pro \cite{comanici2025gemini}), 10 open-source general VLMs (Qwen2.5-VL series \cite{bai2025qwen2}, Qwen3-VL series \cite{Qwen3-VL} and InternVL3 series \cite{zhu2025internvl3}) and 4 open-source embodied VLMs (RoboBrain2.0 series \cite{team2025robobrain} and Cosmos-Reason1-7B \cite{azzolini2025cosmos}). 

\textbf{Evaluation Metrics.} For multiple-choice questions, we use the percentage of correctly answered samples as the accuracy metric. For open-ended questions, we employ the GPT-4o-based evaluator to comprehensively assess the quality of the generated output. This evaluator scores the output across three dimensions: semantic similarity, content completeness, and functional equivalence. Finally, we average the scores from these dimensions to calculate a total score, thereby providing an overall evaluation of the output quality. The evaluation details are available in the appendix. 

\begin{table}[htbp]
  \centering
  \small
  \renewcommand{\arraystretch}{0.9}
  \caption{Comparison of overall model performance on ViFailback-Bench. All metrics are reported as accuracy (\%). \textbf{Bold} scores indicate the best performance, and \underline{underlined} scores indicate the second best performance.}
  \label{tab:bench_overall}
  \begin{tabular}{lrrr}
    \toprule
    \textbf{Model} & \textbf{Lite} & \textbf{Hard} & \textbf{Average} \\
    \midrule
    \rowcolor{gray!20}
    \multicolumn{4}{l}{\textit{General Open-Source Models}} \\
    Qwen2.5-VL-3B-Instruct \cite{bai2025qwen2} & 38.10 & 22.10 & 30.81 \\
    Qwen2.5-VL-7B-Instruct \cite{bai2025qwen2} & 42.41 & 19.26 & 31.87 \\
    Qwen2.5-VL-32B-Instruct \cite{bai2025qwen2} & 46.30 & 32.50 & 40.02 \\
    Qwen2.5-VL-72B-Instruct \cite{bai2025qwen2} & \underline{50.61} & \underline{36.56} & 44.21 \\
    Qwen3-VL-2B-Instruct \cite{Qwen3-VL} & 35.16 & 20.28 & 28.39 \\
    Qwen3-VL-4B-Instruct \cite{Qwen3-VL} & 41.11 & 33.37 & 37.59 \\
    Qwen3-VL-8B-Instruct \cite{Qwen3-VL} & 38.33 & 33.04 & 35.92 \\
    Qwen3-VL-32B-Instruct \cite{Qwen3-VL} & 47.79 & 35.23 & 42.07 \\
    InternVL3-8B \cite{zhu2025internvl3} & 36.48 & 29.82 & 33.45 \\
    InternVL3-78B \cite{zhu2025internvl3} & 42.81 & 30.77 & 37.33 \\
    \midrule
    \rowcolor{gray!20}
    \multicolumn{4}{l}{\textit{Embodied Models}} \\
    RoboBrain2.0-3B \cite{team2025robobrain} & 40.39 & 21.21 & 31.65 \\
    RoboBrain2.0-7B \cite{team2025robobrain} & 40.62 & 19.15 & 30.84 \\
    RoboBrain2.0-32B \cite{team2025robobrain} & 49.92 & 29.22 & 40.50 \\
    Cosmos-Reason1-7B \cite{azzolini2025cosmos} & 38.06 & 28.60 & 33.75 \\
    \midrule
    \rowcolor{gray!20}
    \multicolumn{4}{l}{\textit{General Closed-Source Models}} \\
    GPT-4o \cite{hurst2024gpt} & 48.21 & \textbf{40.00} & \underline{44.47} \\
    Gemini-2.5-Pro \cite{comanici2025gemini} & \textbf{54.64} & 32.45 & \textbf{44.54} \\
    \bottomrule
  \end{tabular}
\end{table}

\begin{table*}[t]
\centering
\renewcommand{\arraystretch}{0.92}
% 设定列间距稍小一点，防止表1太宽溢出
\setlength{\tabcolsep}{2pt}
% 统一使用 footnotesize (或者 small)，保证两个表字体一致
\footnotesize
\caption{Comparison of model performance on ViFailback-Bench \textbf{Lite}. All metrics are reported as accuracy (\%). \textbf{Bold} scores indicate the best performance, and \underline{underlined} scores indicate the second best performance.}
\label{tab:ViFailback-Bench-Lite}
% 使用 tabular* 配合 @{\extracolsep{\fill}} 让表格自动撑满 textwidth
\begin{tabular*}{\textwidth}{@{\extracolsep{\fill}}l*{7}{c}}
\toprule
\textbf{Model} & \textbf{Failure} & \textbf{Failure Keyframe} & \textbf{Failure Subtask} & \textbf{Failure Type} & \textbf{Low-level} & \textbf{Low-level} & \textbf{Average} \\
 & \textbf{Detection} & \textbf{Localization} & \textbf{Localization} & \textbf{Identification} & \textbf{Avoidance} & \textbf{Correction} &  \\
\midrule
\rowcolor{gray!20}
\multicolumn{8}{l}{\textit{General Open-Source Models}} \\
Qwen2.5-VL-3B-Instruct \cite{bai2025qwen2} & 81.20 & 23.15 & 17.98 & 20.22 & 32.74 & 48.60 & 38.10 \\
Qwen2.5-VL-7B-Instruct \cite{bai2025qwen2} & 91.40 & 30.34 & 13.26 & 32.13 & 35.03 & 45.80 & 42.41 \\
Qwen2.5-VL-32B-Instruct \cite{bai2025qwen2} & 74.20 & 26.74 & 11.69 & 66.52 & 44.42 & 51.15 & 46.30 \\
Qwen2.5-VL-72B-Instruct \cite{bai2025qwen2} & 88.80 & 26.07 & 28.76 & 49.89 & 44.42 & \underline{61.58} & 50.61 \\
Qwen3-VL-2B-Instruct \cite{Qwen3-VL} & 55.20 & 29.21 & 12.81 & 43.37 & 29.19 & 38.42 & 35.16 \\
Qwen3-VL-4B-Instruct \cite{Qwen3-VL} & 92.20 & 32.81 & 20.22 & 43.60 & 27.41 & 20.10 & 41.11 \\
Qwen3-VL-8B-Instruct \cite{Qwen3-VL} & 90.00 & 31.24 & 10.34 & 37.08 & 31.98 & 20.10 & 38.33 \\
Qwen3-VL-32B-Instruct \cite{Qwen3-VL} & \underline{93.60} & 36.18 & 26.07 & 48.99 & 38.83 & 34.86 & 47.79 \\
InternVL3-8B \cite{zhu2025internvl3} & 63.00 & 27.39 & 12.56 & 39.95 & 34.11 & 38.60 & 36.48 \\
InternVL3-78B \cite{zhu2025internvl3}  & 75.40 & 27.64 & 29.40 & 38.94 & 39.36 & 41.52 & 42.81 \\
\midrule
\rowcolor{gray!20}
\multicolumn{8}{l}{\textit{Embodied Models}} \\
RoboBrain2.0-3B \cite{team2025robobrain} & 85.40 & 29.89 & 12.58 & 19.10 & 30.46 & 60.56 & 40.39 \\
RoboBrain2.0-7B \cite{team2025robobrain} & 90.20 & 32.13 & 8.31 & 17.98 & 36.80 & 53.18 & 40.62 \\
RoboBrain2.0-32B \cite{team2025robobrain} & 84.80 & 38.43 & 11.46 & \underline{68.31} & 37.82 & 53.44 & 49.92 \\
Cosmos-Reason1-7B \cite{azzolini2025cosmos} & 78.40 & 20.67 & 14.61 & 37.75 & 32.74 & 38.68 & 38.06 \\
\midrule
\rowcolor{gray!20}
\multicolumn{8}{l}{\textit{General Closed-Source Models}} \\
GPT-4o \cite{hurst2024gpt} & 93.40 & 46.97 & 13.93 & 40.90 & 44.16 & 43.26 & 48.21 \\
Gemini-2.5-Pro \cite{comanici2025gemini} & 93.13 & \underline{47.64} & \underline{33.48} & 40.73 & \underline{58.10} & 49.87 & \underline{54.64} \\
\midrule
\textbf{ViFailback-8B (Ours)} & \textbf{98.20} & \textbf{92.58} & \textbf{93.48} & \textbf{90.79} & \textbf{93.15} & \textbf{95.93} & \textbf{93.70} \\
\bottomrule
\end{tabular*}
\end{table*}

\textbf{Finetuning with the ViFailback dataset.} We construct a training split from the ViFailback dataset and use LoRA \cite{hu2022lora} to finetune the Qwen3-VL-8B model for 1 epoch, yielding the model ViFailback-8B.

\begin{table*}[t]
\centering
\renewcommand{\arraystretch}{0.92}
% 保持与上表一致的设置
\setlength{\tabcolsep}{2pt} 
\footnotesize 
\caption{Comparison of model performance on ViFailback-Bench \textbf{Hard}. All metrics are reported as accuracy (\%). \textbf{Bold} scores indicate the best performance, and \underline{underlined} scores indicate the second best performance.}
\label{tab:ViFailback-Bench-Hard}
% 使用 tabular* 配合 @{\extracolsep{\fill}}
\begin{tabular*}{\textwidth}{@{\extracolsep{\fill}}l*{6}{c}}
\toprule
\textbf{Model} & \textbf{Low-level} & \textbf{Low-level} & \textbf{Failure} & \textbf{High-level} & \textbf{High-level} & \textbf{Average} \\
 & \textbf{Avoidance (CoT)} & \textbf{Correction (CoT)} & \textbf{Reason}  & \textbf{Avoidance} & \textbf{Correction} &  \\
\midrule
\rowcolor{gray!20}
\multicolumn{7}{l}{\textit{General Open-Source Models}} \\
Qwen2.5-VL-3B-Instruct \cite{bai2025qwen2} & 3.62 & 7.88 & 31.16 & 31.94 & 37.09 & 22.10 \\
Qwen2.5-VL-7B-Instruct \cite{bai2025qwen2} & 12.77 & 13.64 & 33.10 & 18.95 & 18.28 & 19.26 \\
Qwen2.5-VL-32B-Instruct \cite{bai2025qwen2} & 9.91 & 9.67 & 51.92 & 44.22 & 48.47 & 32.50 \\
Qwen2.5-VL-72B-Instruct \cite{bai2025qwen2} & 13.05 & 18.04 & 54.26 & 47.78 & 51.22 & 36.56 \\
Qwen3-VL-2B-Instruct \cite{Qwen3-VL} & 0.00 & 2.63 & 31.16 & 31.94 & 37.09 & 20.28 \\
Qwen3-VL-4B-Instruct \cite{Qwen3-VL} & 11.23 & 13.94 & 57.01 & 44.83 & 41.37 & 33.37 \\
Qwen3-VL-8B-Instruct \cite{Qwen3-VL} & 12.77 & 16.45 & 51.98 & 43.47 & 41.89 & 33.04 \\
Qwen3-VL-32B-Instruct \cite{Qwen3-VL} & 14.11 & 19.34 & \underline{60.53} & 40.44 & 43.09 & 35.23 \\
InternVL3-8B \cite{zhu2025internvl3} & 5.23 & 5.51 & 44.01 & 47.32 & 48.85 & 29.82 \\
InternVL3-78B \cite{zhu2025internvl3} & 8.39 & 11.45 & 39.30 & 46.64 & 49.63 & 30.77 \\
\midrule
\rowcolor{gray!20}
\multicolumn{7}{l}{\textit{Embodied Models}} \\
RoboBrain2.0-3B \cite{team2025robobrain} & 6.47 & 2.68 & 40.81 & 27.19 & 30.11 & 21.21 \\
RoboBrain2.0-7B \cite{team2025robobrain} & 4.80 & 1.18 & 35.07 & 26.02 & 29.89 & 19.15 \\
RoboBrain2.0-32B \cite{team2025robobrain} & 3.55 & 8.57 & 52.36 & 38.37 & 44.95 & 29.22 \\
Cosmos-Reason1-7B \cite{azzolini2025cosmos} & 9.20 & 8.30 & 36.27 & 44.02 & 46.69 & 28.60 \\
\midrule
\rowcolor{gray!20}
\multicolumn{7}{l}{\textit{General Closed-Source Models}} \\
GPT-4o \cite{hurst2024gpt} & \underline{18.93} & 18.86 & 59.28 & \underline{49.53} & \underline{54.96} & \underline{40.00} \\
Gemini-2.5-Pro \cite{comanici2025gemini} & 13.04 & \underline{26.90} & 53.74 & 21.85 & 47.62 & 32.45 \\
\midrule
\textbf{ViFailback-8B (Ours)} & \textbf{47.95} & \textbf{65.33} & \textbf{83.97} & \textbf{85.36} & \textbf{81.79} & \textbf{72.64} \\
\bottomrule
\end{tabular*}
\end{table*}

\subsection{Main Results}
\label{sec:main_results}
\textbf{Overall Model Performance.} As shown in Table~\ref{tab:bench_overall}, the capabilities of all models for robotic manipulation failure diagnosis and correction are far from satisfactory. This performance degradation is even more pronounced in the open-ended setting. The best performance in the closed-ended setting was 54.64\%, achieved by Gemini-2.5-Pro, while GPT-4o achieved the top score of 40.00\% in the open-ended setting.

\textbf{Closed-Ended Performance Analysis.} Results in Table~\ref{tab:ViFailback-Bench-Lite} show the performance comparison on the ViFailback-Bench Lite. While baselines achieve high accuracy on failure detection, they fall short in the crucial tasks of failure localization and low-level guidance. By supervised finetuning on our dataset, ViFailback-8B achieves a 39.14\% performance gain over the best-performing baseline, Gemini-2.5-Pro.

\textbf{Open-Ended Performance Analysis.} As illustrated in Table~\ref{tab:ViFailback-Bench-Hard}, GPT-4o achieves the overall best performance in the baseline. Notably, in the Failure reason task, the open-source Qwen3-VL-32B-Instruct even surpasses the proprietary models, reaching the highest accuracy of 60.53\%. In contrast, ViFailback-8B demonstrates a 39.6\% performance gain over its zero-shot counterpart and significantly outperforms all baseline models. Furthermore, by comparing Table~\ref{tab:ViFailback-Bench-Lite} and Table~\ref{tab:ViFailback-Bench-Hard}, we also observe that when multi-step reasoning is required to localize failures and determine low-level corrective actions, the performance of all models drops substantially—indicating that current VLMs struggle to generate step-by-step guidance (from failure detection and localization to guiding avoidance or correction) for robotic manipulation failures.

\begin{figure*}[htbp]
    \centering
    \includegraphics[width=0.86\textwidth]{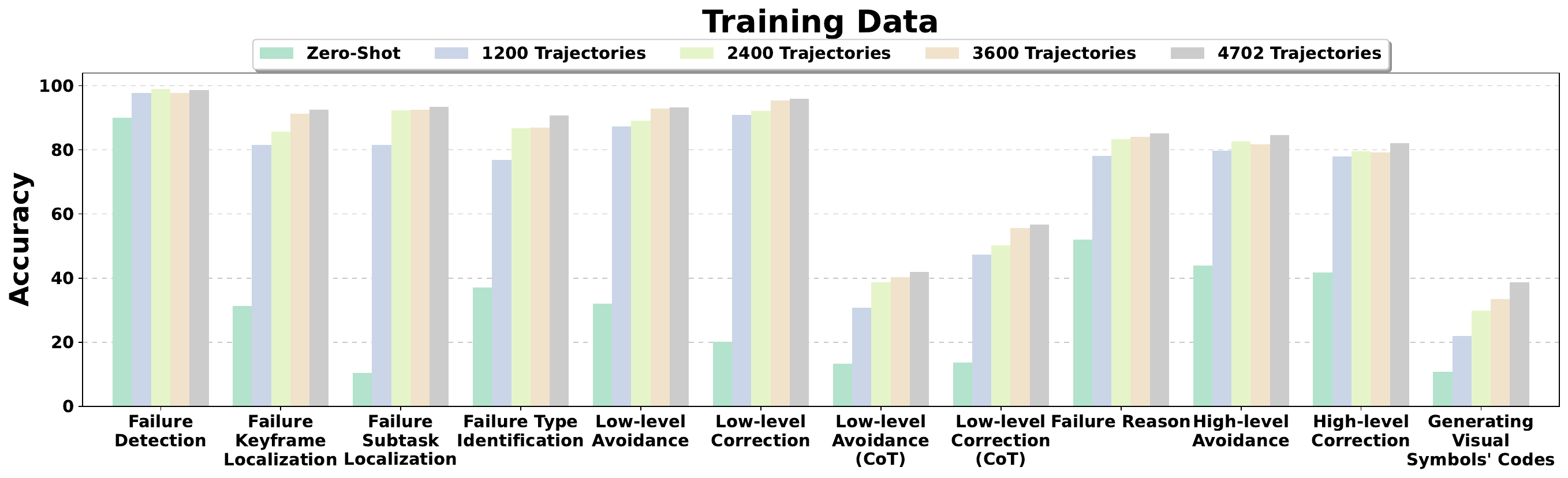}
    \caption{\textbf{Finetuning results}. Qwen3-VL-8B with increasing training data of the ViFailback dataset from 1,200 trajectories to 4,702 trajectories' corresponding VQA pairs. Accuracy improvements are observed across most VQA types as the training data increases. All metrics are reported as accuracy (\%)}
    \label{fig:scale}
\end{figure*}

\textbf{Performance Scales with Data Size.} We construct the training split from the ViFailback dataset. Our full training set comprises 52,418 VQA pairs generated from 4,702 trajectories, covering 95 distinct tasks. To analyze the impact of data scaling, we create and evaluate training subsets of increasing sizes, specifically using data from 1,200, 2,400, 3,600, and the full 4,702 trajectories. As depicted in Figure \ref{fig:scale}, increasing the training data volume generally leads to improved model performance across most VQA types. Significant gains are observed in the ViFailback-Bench Lite tasks, even when finetuning with limited data (\eg, 1,200 trajectories). This may be because the tasks are related to the VLM's basic capabilities, but the VQA formats of ViFailback-Bench Hard are rare in the base models' pre-training data. Furthermore, we examine the models' ability in visual symbol generation. The results of ``Generating Visual Symbols' Codes'' in Figure \ref{fig:scale} show performance on this task scales consistently with the training data volume, reaching 38.73\%. In terms of annotation efficiency, relative to the zero-shot baseline, each additional person-hour of annotation yields an average 0.64\% accuracy improvement. Crucially, this performance trend has not yet reached saturation, indicating that further data expansion could yield additional improvements.

\subsection{Application in Real-world Manipulation}
\textbf{Real-world Experiment Setup.} To further demonstrate the effectiveness of ViFailback-8B in enhancing the ability of VLA models to handle failures in downstream robotic manipulation tasks, we integrate it with a leading VLA model, $\pi _{0.5}$ \cite{black2025pi_}, into a unified control system. The system facilitates failure recovery by providing $\pi_{0.5}$ with corrective visual and textual guidance from ViFailback-8B. We conduct experiments on the ALOHA dual-arm robot platform, equipped with two wrist cameras and one head camera. The $\pi _{0.5}$ model takes observations from all three cameras as input, while ViFailback-8B uses only the head camera's observations for failure diagnosis and correction. Given that the original $\pi_{0.5}$ model lacks a native capability to directly follow visual symbols, we devised two distinct experimental setups to address this limitation:

\begin{itemize}[leftmargin=2em, topsep=0pt, itemsep=0pt]
    \item \textbf{Incorporating Visual Symbols-Following Dataset (VSF Method).} We constructed a visual symbols-following dataset by collecting low-level motion trajectories (\eg, move the left gripper to the left) and annotating them with visual symbols. Inspired by PEEK \cite{zhang2025peek}, we mask out irrelevant regions in the head camera view and the entire observation from the wrist camera, which receives no guidance. Then the visual symbols-following data are used to finetune the $\pi_{0.5}$ together with the task-specific expert demonstrations to enable the $\pi_{0.5}$ to end-to-end follow the visual symbols.
    \item \textbf{Point-based Motion Control (PMC Method).} We employ a low-level motion controller to drive the end-effector to follow the visual symbols. For example, when a movement is required, the end-effector first moves to the target point indicated by the visual symbol. If grasping is needed, we further use GraspNet \cite{mousavian20196} to estimate the appropriate grasp pose.

\end{itemize}
For each task, we collect 20 expert demonstrations for finetuning. To analyze the potential impact of visual symbols-following data on models trained with original demonstrations, the baseline experiments use two versions of finetuned $\pi _{0.5}$ models: one is finetuned with only expert demonstrations (base data), and the other is finetuned with both base data and visual symbols-following data (symbol data).

\textbf{Workflow of the Policy Correction System.} The ALOHA robot first attempts to complete the manipulation task using the finetuned $\pi _{0.5}$ model under the normal textual task prompt, and our ViFailback-8B simultaneously listens to the robot's video stream at a set interval. If ViFailback-8B detects a failure, it generates a CoT-based diagnosis and low-level guidance, especially the code for drawing visual symbols based on the current observation of the robot. The visual symbols are then overlaid on the robot's camera view and fed into the finetuned $\pi _{0.5}$ model along with the textual prompt of the correction guidance to guide the robot to recover from the failure. 

\begin{table}[t]
\centering
\renewcommand{\arraystretch}{0.9}
\caption{Success rates of different methods on three tasks.}
\label{tab:task_comparison}
\resizebox{\columnwidth}{!}{
\begin{tabular}{lcccc}
\toprule
\textbf{Method} & \textbf{PlaceOne} & \textbf{PlaceTwo} & \textbf{Pull\&Place} & \textbf{Average} \\
\midrule
\multicolumn{5}{l}{\textbf{w/o ViFailback Correction}} \\ 
$\pi_{0.5}$ (base \& symbol) & 14/21 & 9/21 & 10/21 & 52.4\% \\
$\pi_{0.5}$ (base) & 13/21 & 9/21 & 10/21 & 50.8\% \\ 
\midrule
\multicolumn{5}{l}{\textbf{w/ ViFailback Correction}} \\ 
$\pi_{0.5}$ (base \& symbol) + VSF  & 18/21 & 13/21 & \textbf{15/21} & 73.0\% \\ 
$\pi_{0.5}$ (base) + PMC& \textbf{19/21} & \textbf{16/21} & 12/21 & \textbf{74.6\%} \\
\bottomrule
\end{tabular}
}
\vspace{-0.5em}
\end{table}

\begin{figure}[t]
    \centering
    \includegraphics[width=0.9\columnwidth]{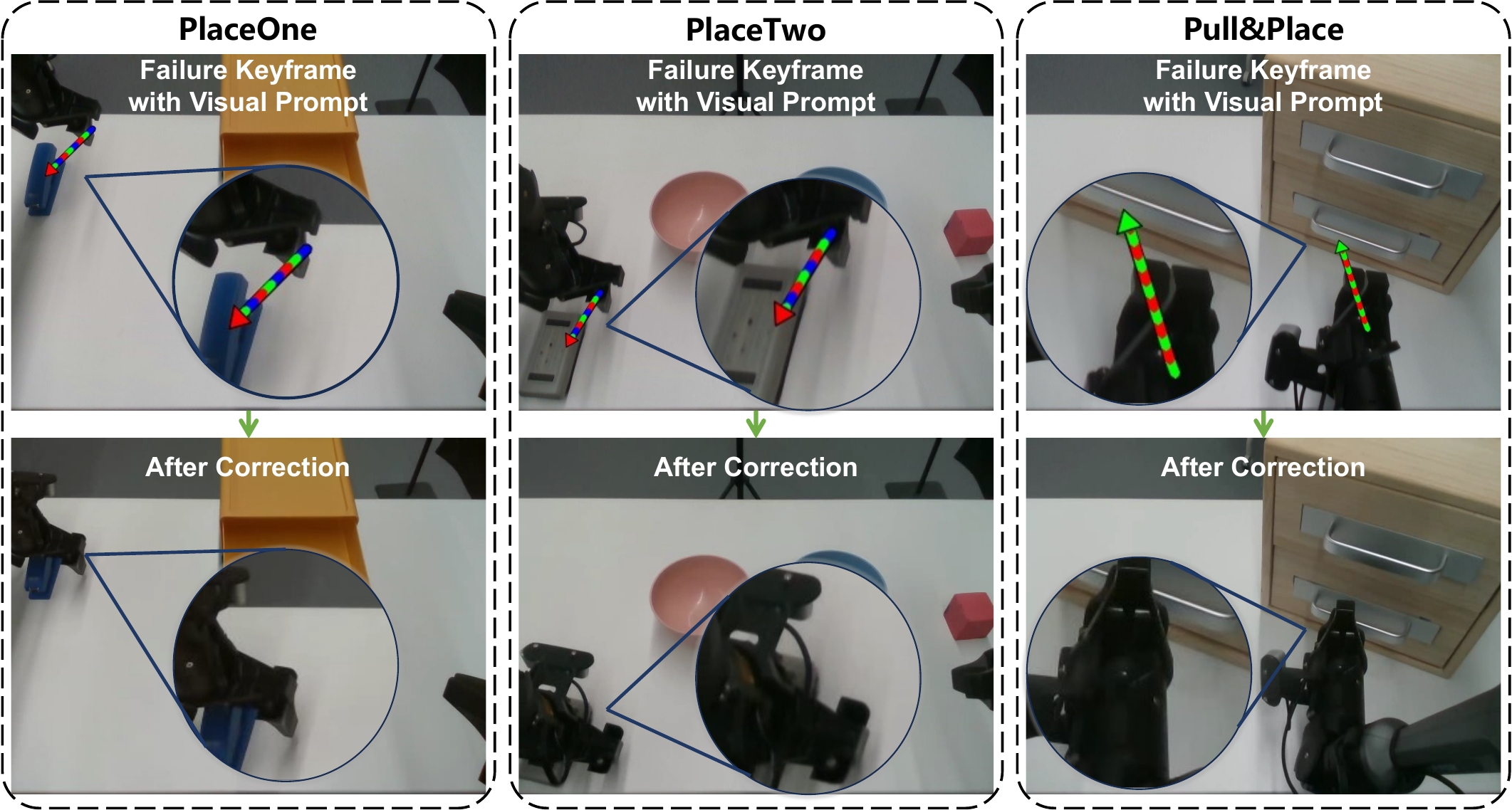}
    \caption{Under the visual symbols' guidance generated by ViFailback-8B in real-time, the robot successfully recovers from failures in three representative manipulation tasks: PlaceOne, PlaceTwo, and Pull\&Place in the real-world experiments.}
    \label{fig:experiment}
\end{figure}

\textbf{Real-world Experiment Results.} 
We conduct experiments on three representative manipulation tasks: \textbf{PlaceOne}: ``\textit{Use the left arm to place the stapler into the top drawer.}'', \textbf{PlaceTwo}: ``\textit{Use the left arm to put the eraser into the pink bowl. Then use the right arm to put the red cube into the blue bowl.}'', and \textbf{Pull\&Place}: ``\textit{Use the right arm to pull the bottom drawer open. Then use the left arm to place the blue cube into the drawer.}''. All three tasks are unseen in the ViFailback dataset. To ensure a fair comparison, we first randomly initialize the objects' positions and ensure the same initial layout and setup for the test of each method. For each task, we conduct 21 trials and record the success rates of each method, as shown in Table \ref{tab:task_comparison}. Comparing the success rates of the two baseline methods that do not use ViFailback-8B, we observe no significant difference, indicating that the visual symbols-following data does not affect the performance of the $\pi_{0.5}$ model executing the original tasks. When the ViFailback-8B is integrated into the system, the success rates of the $\pi_{0.5}$ model with the VSF method and PMC method are significantly improved—by 20.6\% and 23.8\%, respectively—indicating that the ViFailback-8B can effectively assist the VLA model in recovering from failures in real-world manipulation tasks. Notably, PMC outperforms VSF in two tasks, which may be due to the lack of visual symbols-following data for the $\pi_{0.5}$ model to learn the low-level guidance action following. 

%% file: sec/5_conclusion.tex
\section{Discussion}
In this work, we focus on learning from failure videos to enable diagnosis and correction. We believe that correcting VLA failures based on visual symbols is a more direct and effective approach than language instructions, and critically, it provides the actionable supervisory signal to bridge real-time correction with policy learning. While our framework provides this learning signal from videos, the action distribution of failure trajectories also contains a wealth of valuable information that needs to be utilized in future work.

\section{Conclusion}
In this paper, we introduced ViFailback, a novel framework centered on intuitive visual symbols to diagnose, correct, and ultimately learn from robotic manipulation failures. These symbols serve as both a high-efficiency annotation tool and a direct, interpretable mechanism for corrective guidance. Using this methodology, we constructed the ViFailback dataset (58,128 VQA pairs) and the ViFailback-Bench (Lite and Hard). Our finetuned model, ViFailback-8B, not only significantly outperforms state-of-the-art models on our benchmark, but also learns to generate visual symbols itself. We validated this in real-world experiments, where ViFailback-8B's visual guidance significantly improved VLA task success, demonstrating a critical step towards enabling policies to learn from failure.

\section*{Acknowledgements}
This work was supported in part by the
Shanghai Municipal Science and Technology Major Project No.~2025SHZDZX025G14,
Ant Group, and
National Natural Science Foundation of China under Grant No.~U25A20442, 
62306175, 62225302, and 92567303.

%% file: sec/X_suppl.tex
\clearpage
\setcounter{page}{1}
\maketitlesupplementary
\renewcommand{\theHsection}{A\arabic{section}}
\setcounter{section}{0}
\setcounter{figure}{0}
\setcounter{table}{0}

\section{Table of Content}
\label{sec:table}
This supplementary material contains:
\begin{itemize}
\item \cref{sec:vp} \textbf{Design of Visual Symbols} 
\item \cref{sec:annotation} \textbf{Task Design and Data Annotation}
\item \cref{sec:dataset} \textbf{Details of Dataset and Benchmark} 
\item \cref{sec:setting} \textbf{Details of Finetuning and Evaluation} 
\item \cref{sec:real} \textbf{Details of Real-world Experiments}
\end{itemize}

\section{Design of Visual Symbols}
\label{sec:vp}
\begin{figure*}[t]
    \centering
    \includegraphics[width=1\textwidth]{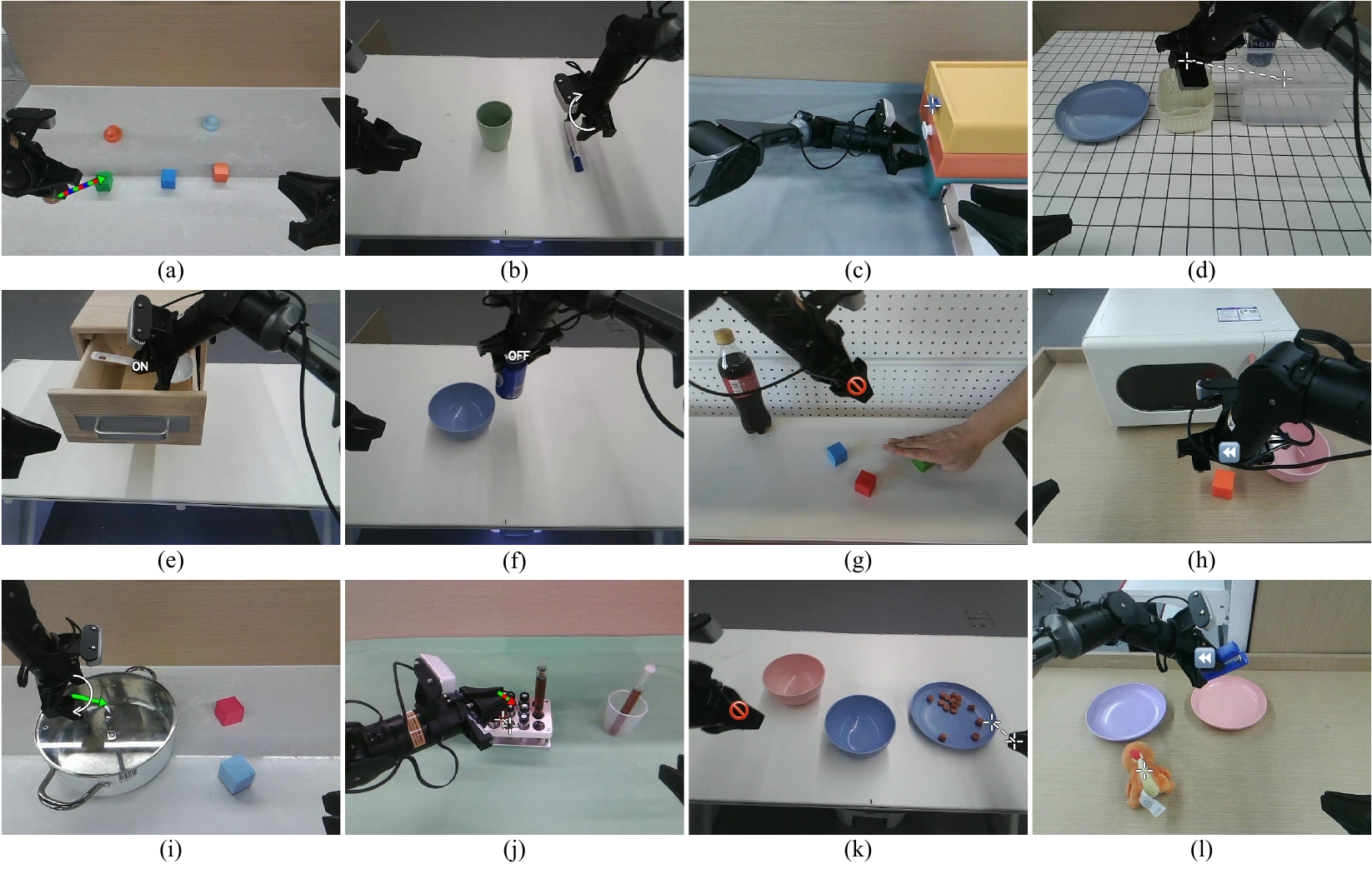}
    \caption{\textbf{Examples of individual visual symbols and their combination.} \textbf{Top and Middle Rows:} Instances of individual visual symbols. \textbf{Bottom Row:} Instances of their usage in combination.}
    \label{fig:vs_symbols_examples}
\end{figure*}

In this section, we present concrete examples to illustrate the application of individual visual symbols, as well as their usage in combination. As shown in Figure \ref{fig:vs_symbols_examples}, the visual symbols in each instance have the following meanings:
\begin{enumerate}[label=(\alph*)]
  \item Move the left gripper to the \textcolor{green}{right} significantly, \textcolor{red}{forward} significantly, and \textcolor{blue}{downward} significantly, aligning the held object with the green cube for placement.
  \item Rotate the right gripper clockwise to achieve the correct grasp pose for the marker.
  \item The correct target (as indicated by the crosshair) is the top drawer handle rather than the bottom one.
  \item The dual crosshairs linked by a dashed line indicates an alignment requirement between the two targeted objects---in this instance, placing the held object into the transparent box.
  \item Open the right gripper, thereby releasing the spatula into the drawer.
  \item Keep the right gripper closed, preventing the coke from slipping during movement.
  \item Hold the left arm still until the human intervention is removed.
  \item Move the right arm back to its initial pose.
  \item Move the left gripper to the \textcolor{green}{right} slightly and rotate it clockwise, achieving the correct grasp pose for the lid.
  \item Move the left gripper to the \textcolor{green}{right} slightly and \textcolor{red}{backward} slightly to align the vial with the specific slot.
  \item Hold the left arm still and align the right gripper with the edge of the blue plate for grasping.
  \item The left arm targets the wrong object, resets to the initial state, and then grasps the correct target---the doll.
\end{enumerate}

\section{Task Design and Data Annotation}
\label{sec:annotation}

\subsection{Task Design Details}

As mentioned in Section \ref{sec:data_collection}, our designed 100 real-world manipulation tasks cover diverse manipulation skills. All 100 tasks and their corresponding trajectory counts are listed in Figure \ref{fig:long_task_table}.

\begin{figure*}[h] 
    \centering
    \includegraphics[width=0.95\linewidth]{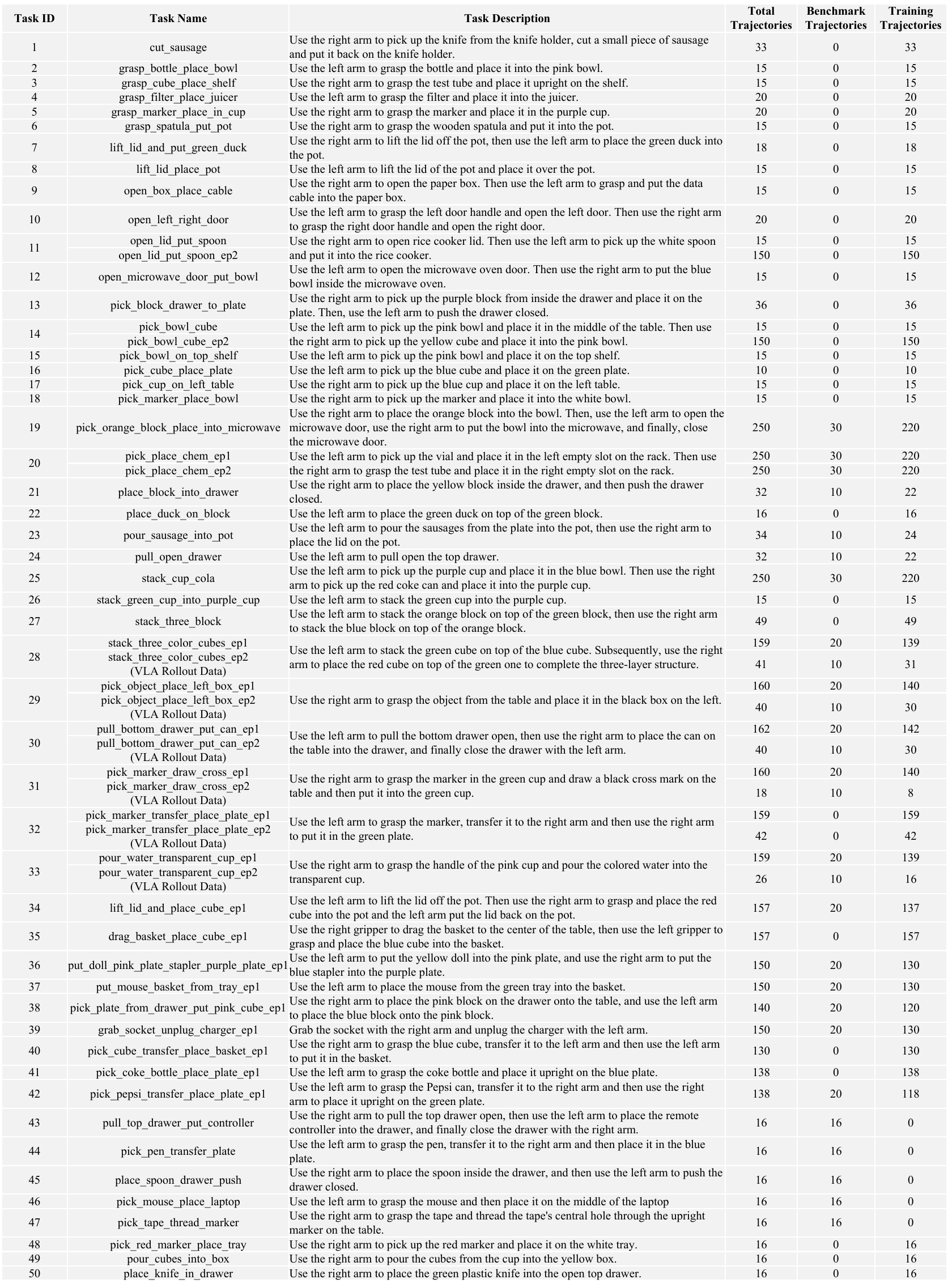}
    \caption{Full list of our designed tasks. (Continued on next page)}
    \label{fig:long_task_table}
\end{figure*}

 % \clearpage 

\begin{figure*}[t]
    \centering
    \ContinuedFloat 
    \includegraphics[width=0.95\linewidth]{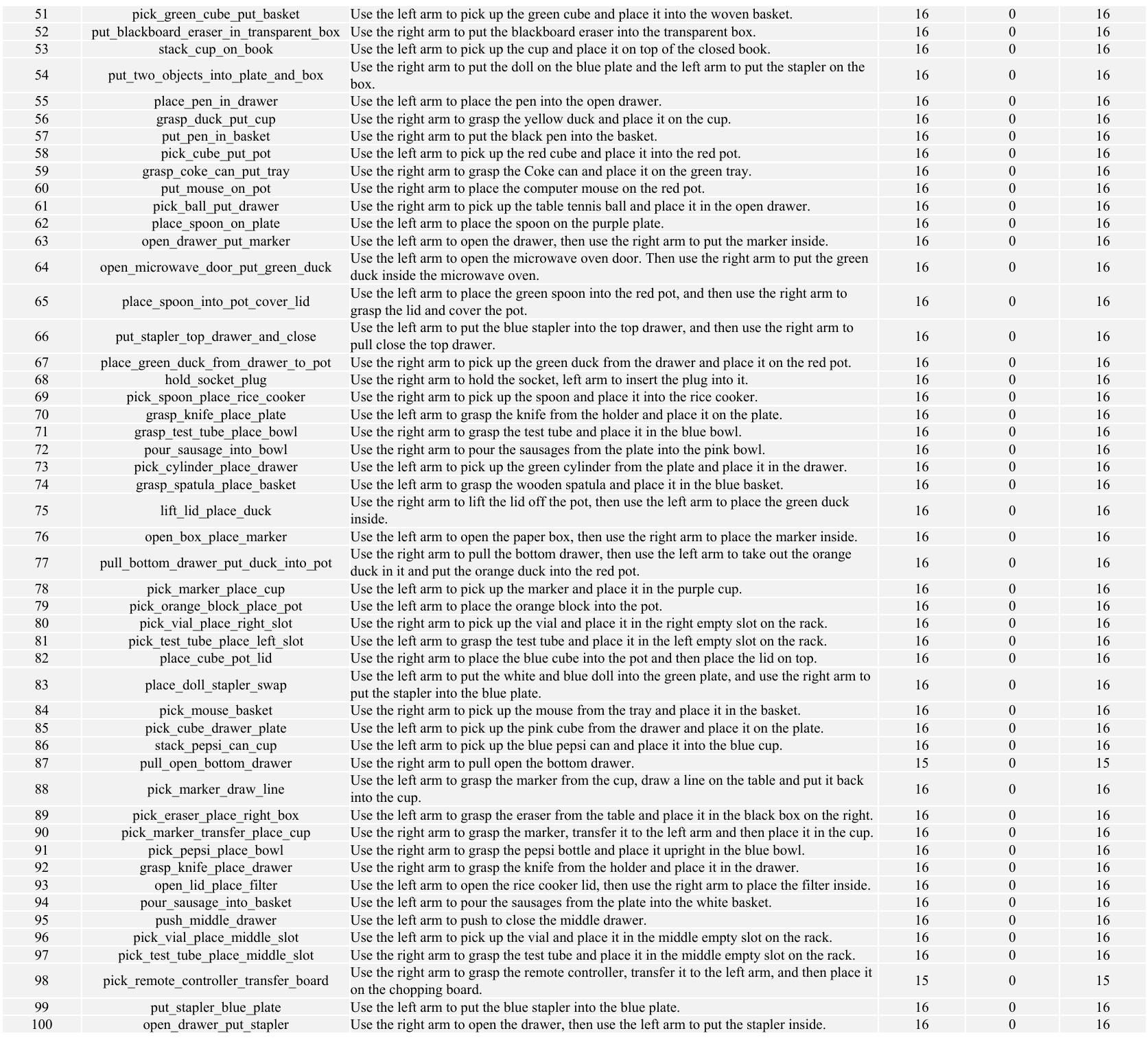}
    \caption{Full list of our designed tasks (continued).}
\end{figure*}

 % \clearpage

\subsection{Data Annotation Tools}
\begin{figure*}[h] 
    \centering
    \includegraphics[width=\linewidth]{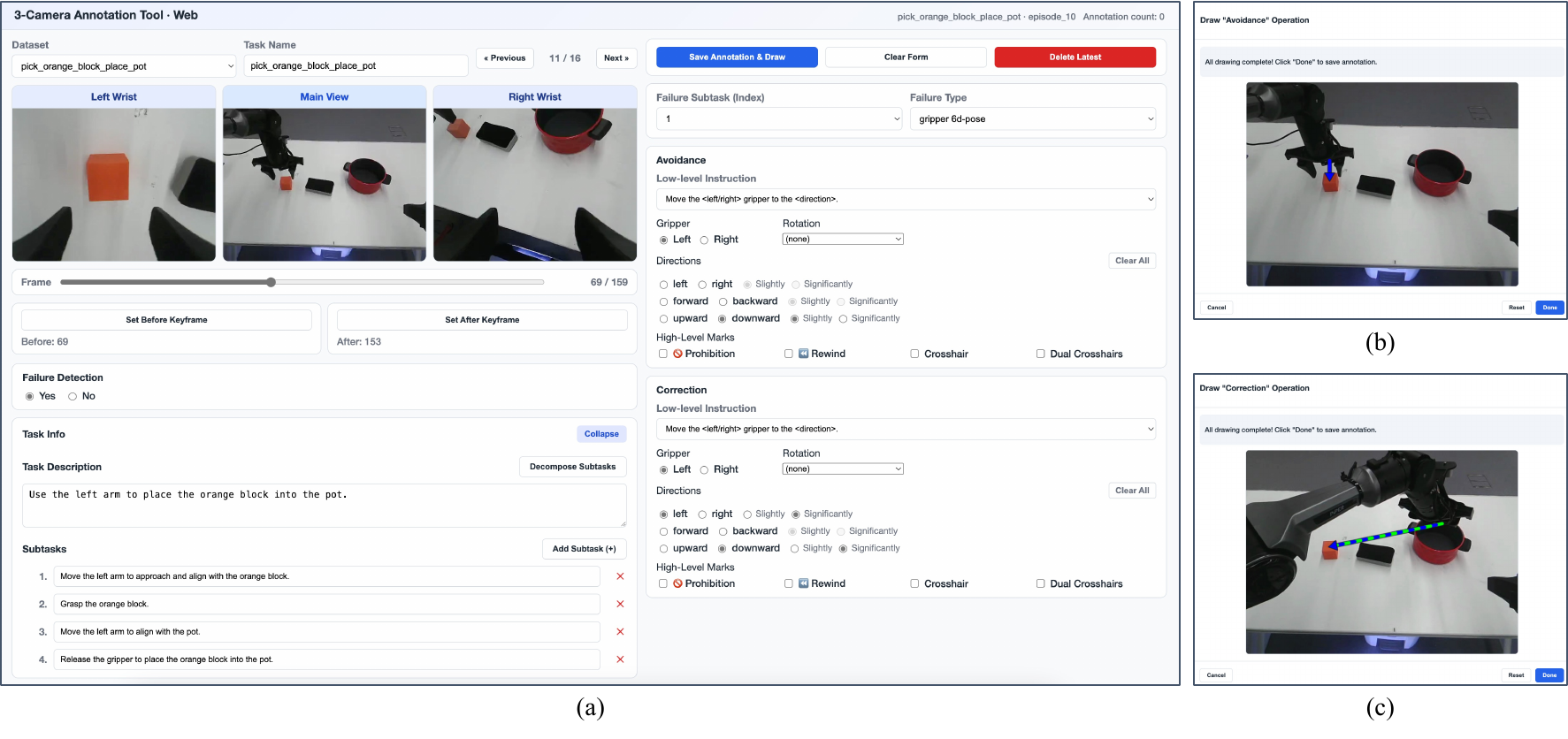}
    \caption{The front-end UI for annotation, including (a) the main annotation interface, (b) the visual symbol drawing interface for avoiding failure, and (c) the visual symbol drawing interface for recovering from failure.}
    \label{fig:tools}
\end{figure*}

The front-end annotation tool is illustrated in Figure \ref{fig:tools}. Annotators can complete the preliminary annotation via click-and-drag mouse operations. Annotations are performed based on the head camera view. Additionally, the wrist camera view (if available) can be loaded to assist annotators in determining spatial positions. Finally, we sample the video at 1 fps and add the selected keyframes to the sampling list.

% \subsection{User Study}

% \section{VQA Design and Examples}
% \label{sec:vqa}

\section{Details of Dataset and Benchmark}
\label{sec:dataset}

\subsection{VQA Design and Examples}

\begin{figure*}[t]
    \centering
    \includegraphics[width=1\textwidth]{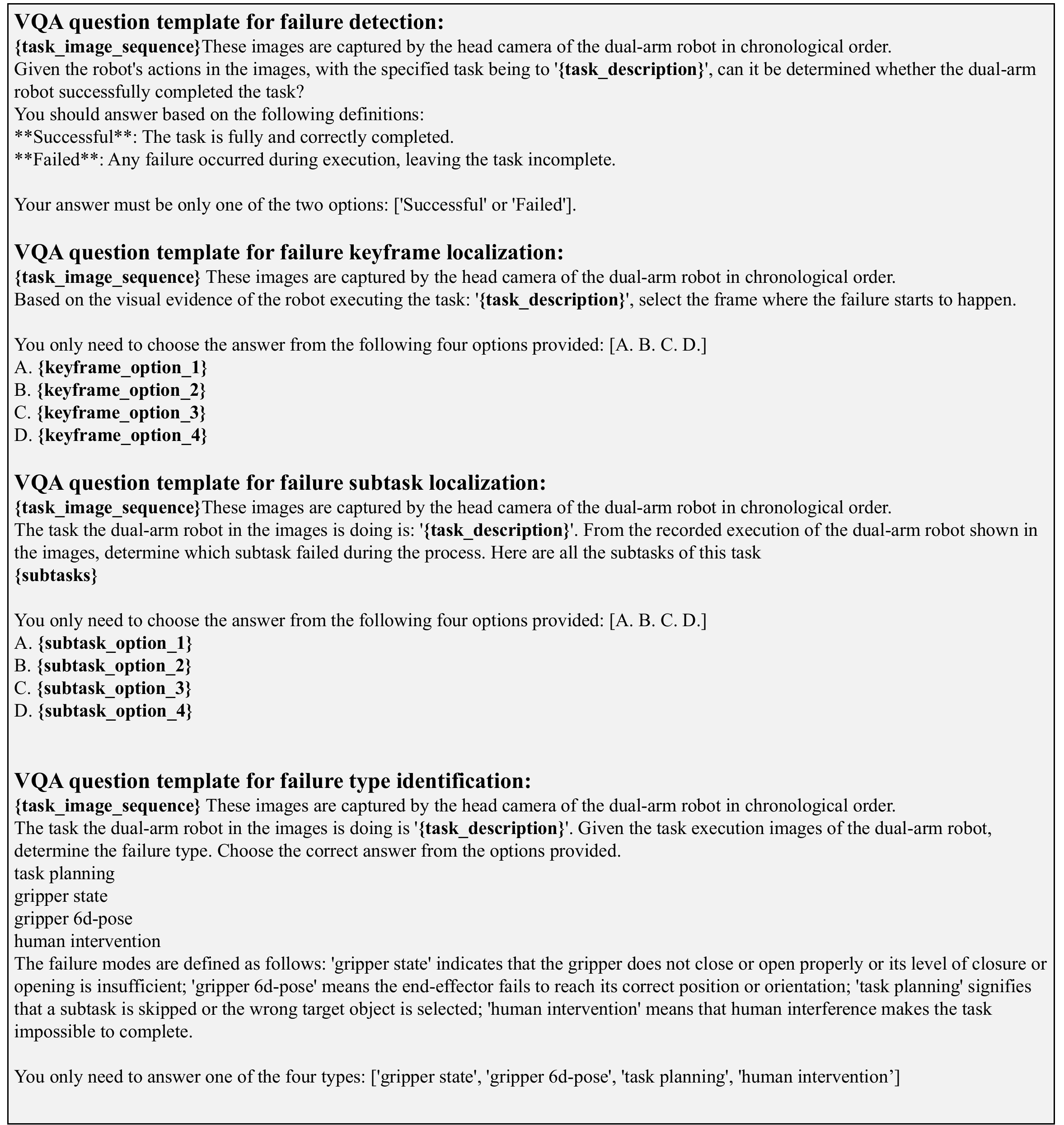}
    \caption{VQA question templates for failure detection, failure keyframe localization, failure subtask localization, and failure type identification.}
    \label{fig:low_level_prompt}
\end{figure*}

\begin{figure*}[t]
    \centering
    \includegraphics[width=1\textwidth]{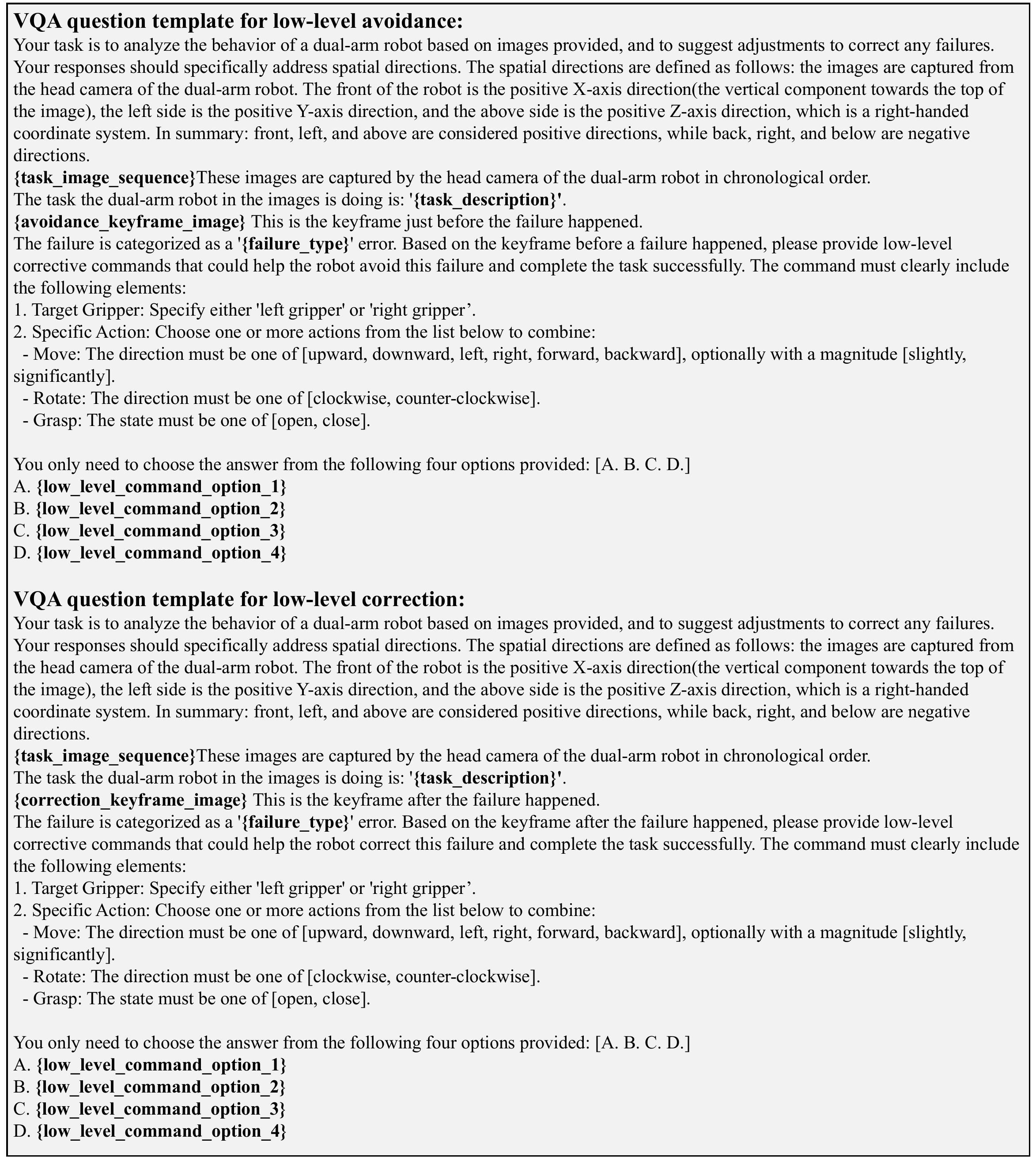}
    \caption{VQA question templates for low-level textual guidance (low-level avoidance and low-level correction).}
    \label{fig:low_level_guidance_prompt}
\end{figure*}

\subsubsection{Design of Closed-ended VQA}
The closed-ended VQA question types include failure detection, failure keyframe localization, failure subtask localization, failure type identification, and low-level textual guidance (avoidance and correction). For all tasks excluding low-level textual guidance, we randomly sample the corresponding annotation pools to construct distractor candidates. The question templates are shown in Figure \ref{fig:low_level_prompt}.

For low-level textual guidance, we construct three challenging distractors using a hybrid sampling strategy. We combine a static pool of common hard-coded actions (\eg, ``Hold still'', ``Open gripper'') with a larger, dynamic pool of all possible commands, ensuring that all sampled distractors are unique and not the ground truth. To increase difficulty, distractors are semantically constrained to match the active gripper (\eg, ``left'' or ``right'') of the correct answer, making all options contextually plausible. The question templates are shown in Figure \ref{fig:low_level_guidance_prompt}.

\begin{figure*}[t]
    \centering
    \includegraphics[width=1\textwidth]{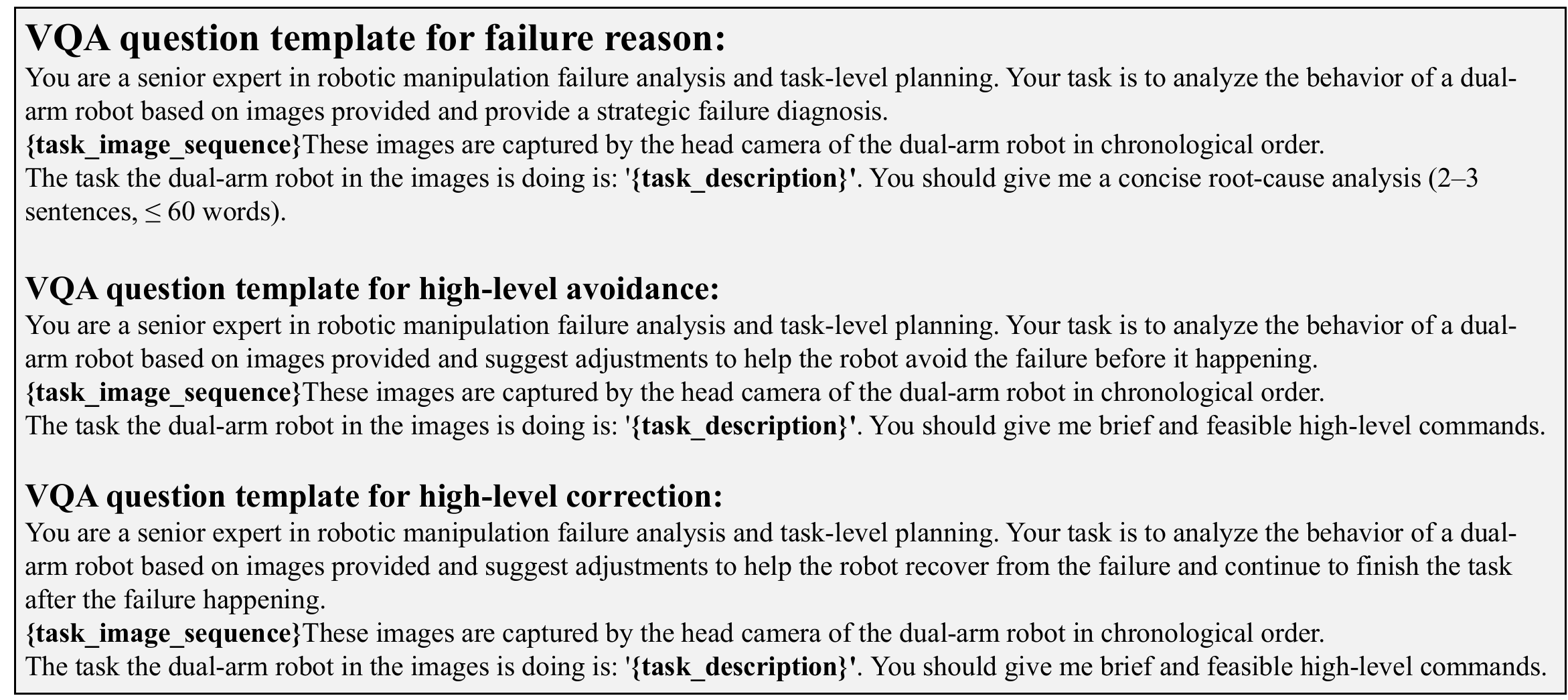}
    \caption{VQA question templates for failure reason and high-level textual guidance (avoidance and correction).}
    \label{fig:high_level_prompt}
\end{figure*}

\begin{figure*}[t]
    \centering
    \includegraphics[width=1\textwidth]{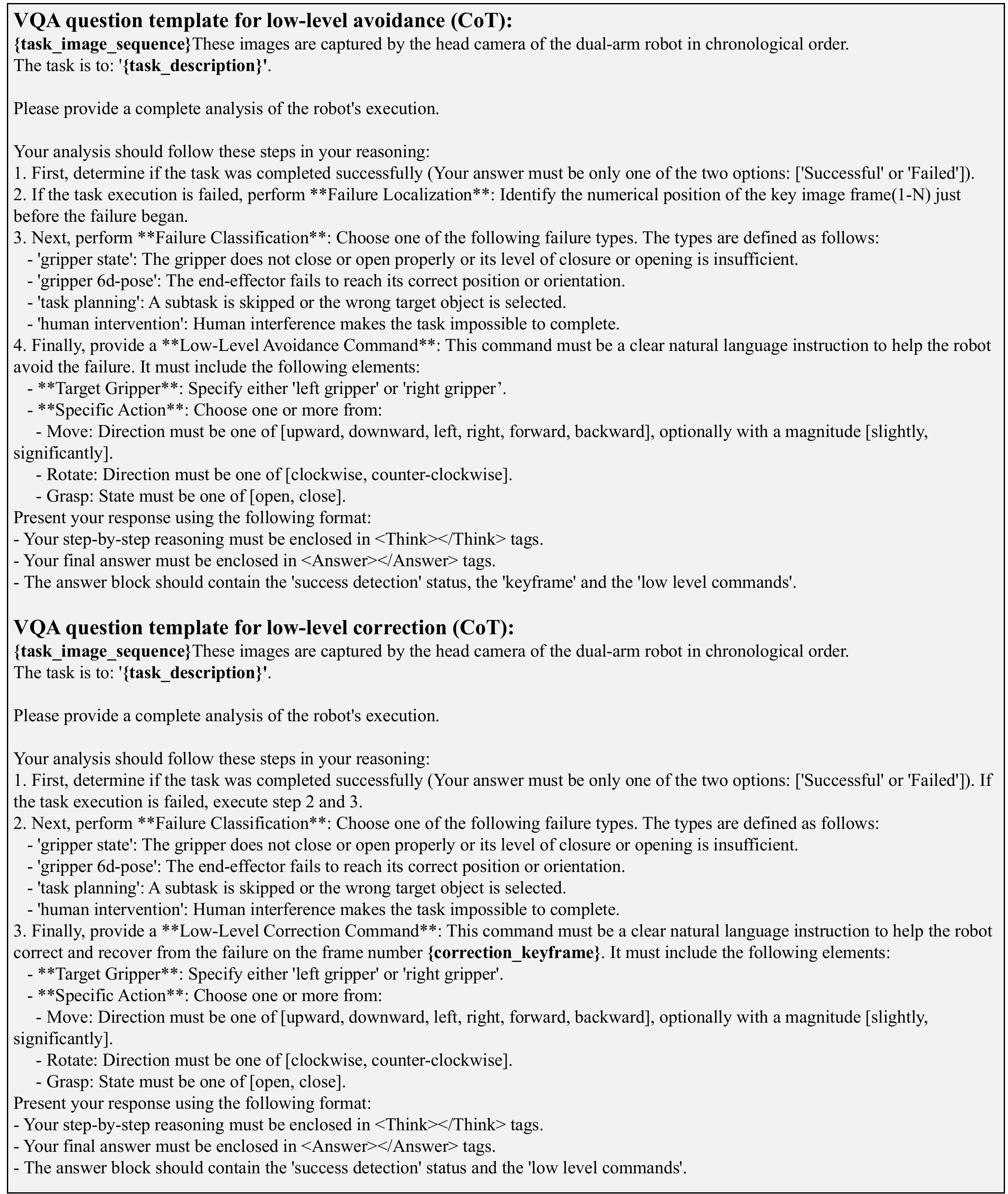}
    \caption{VQA question templates for CoT form of low-level textual guidance (avoidance and correction).}
    \label{fig:low_level_cot_prompt}
\end{figure*}

\subsubsection{Design of Open-ended VQA}
The open-ended VQA question types include failure reason, CoT form of low-level textual guidance (avoidance and correction), and high-level textual guidance (avoidance and correction). The evaluation of open-ended VQA is introduced in Section \ref{sec:eval_prompt}. The question templates are shown in Figure \ref{fig:high_level_prompt} and Figure \ref{fig:low_level_cot_prompt}.

\begin{figure*}[t]
    \centering
    \includegraphics[width=1\textwidth]{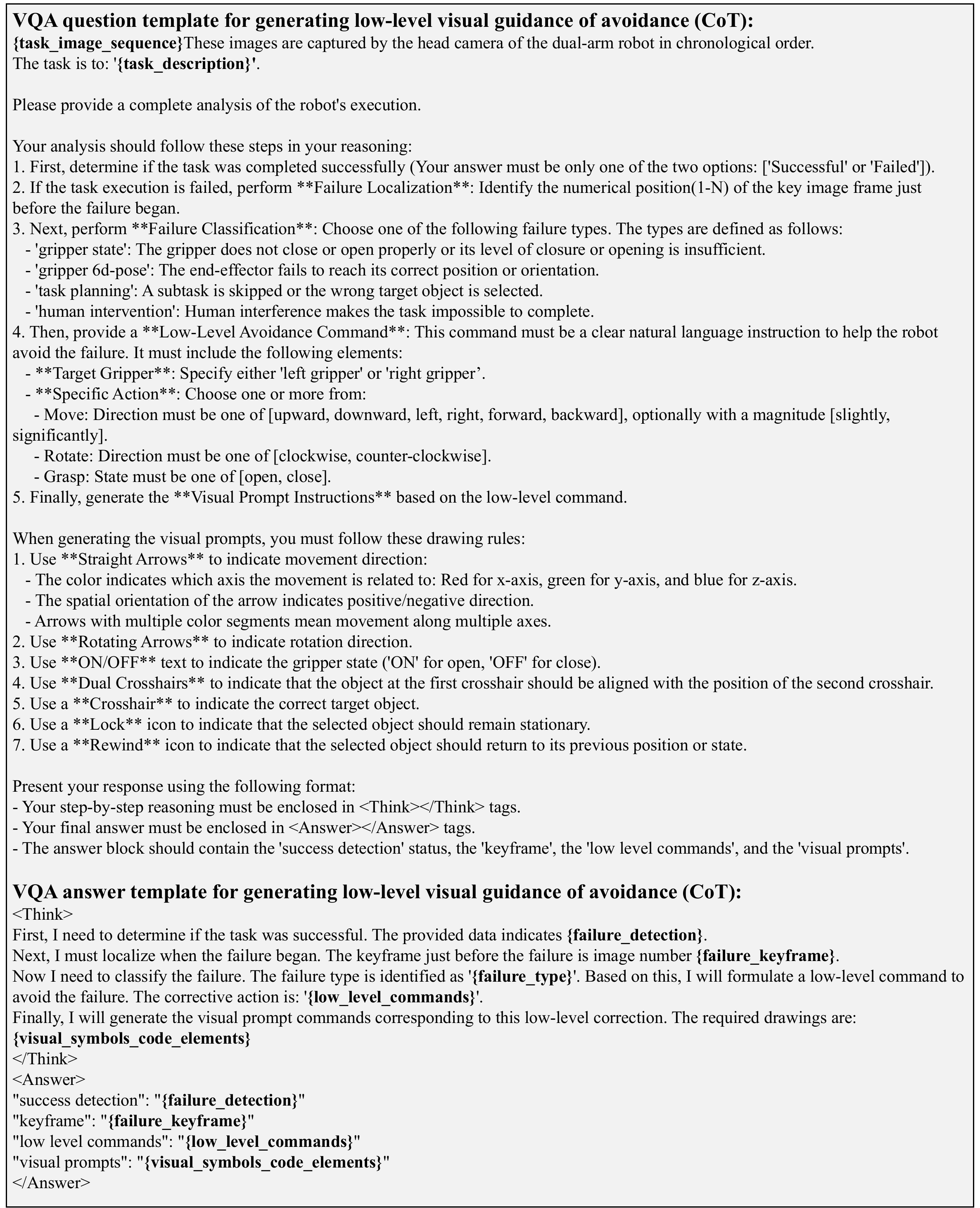}
    \caption{VQA templates for generating low-level visual guidance of avoidance (CoT).}
    \label{fig:cot_code_avoid_prompt}
\end{figure*}

\begin{figure*}[t]
    \centering
    \includegraphics[width=1\textwidth]{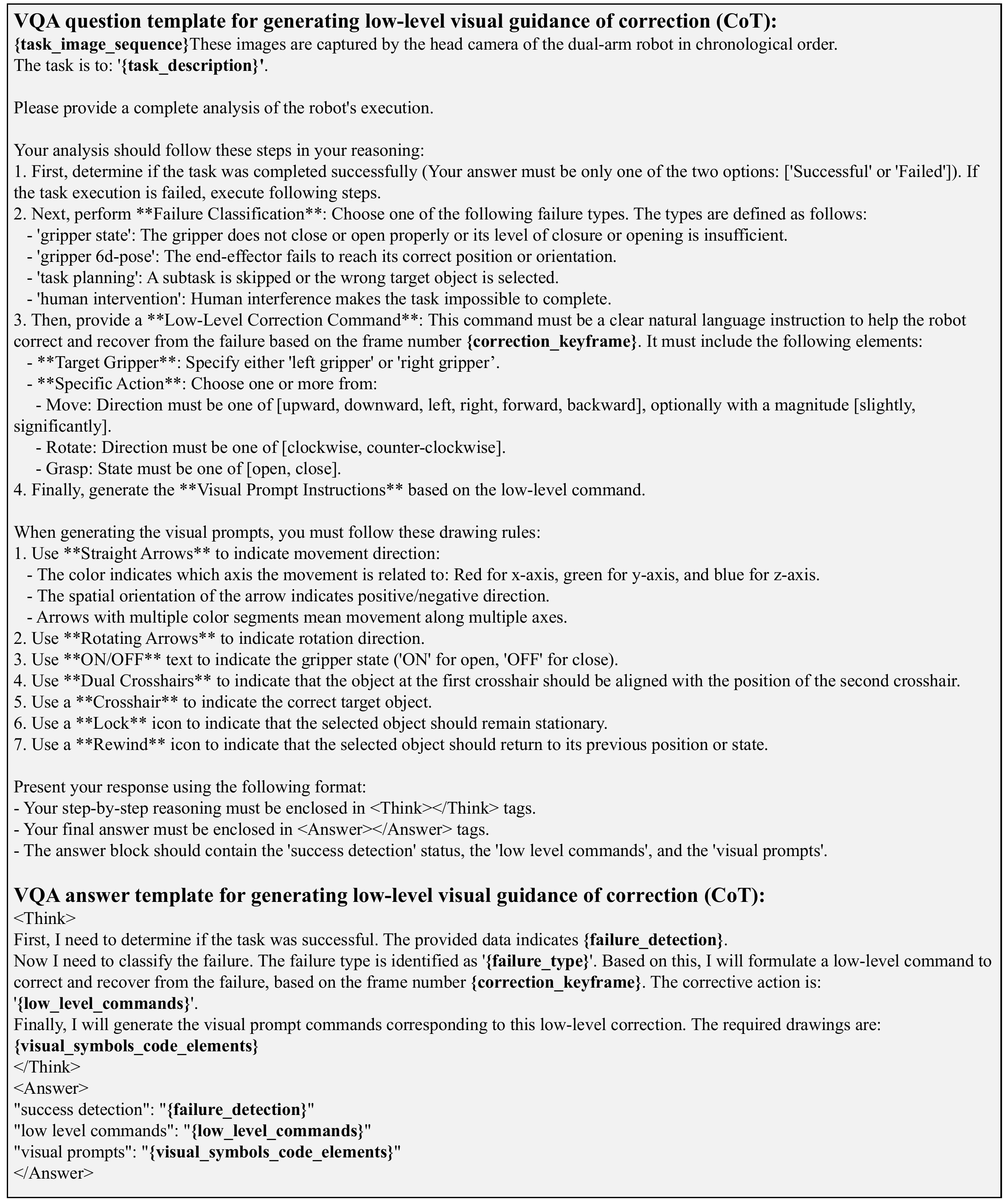}
    \caption{VQA templates for generating low-level visual guidance of correction (CoT).}
    \label{fig:cot_code_correct_prompt}
\end{figure*}

Additionally, there is a special type of VQA called visual guidance, introduced in Section \ref{sec:task_definition} and \ref{sec:main_results}, which is used to train and evaluate our VLM model's ability to generate code elements and draw our designed visual symbols. The question and answer templates are shown in Figure \ref{fig:cot_code_avoid_prompt} and Figure \ref{fig:cot_code_correct_prompt}.

\subsection{Dataset Statistics}
All the VQA pairs in our dataset are generated from 5,202 real-world manipulation trajectories of an ALOHA robot. The distribution of trajectory durations is listed in Table \ref{tab:duration}, which indicates that our trajectory data encompasses both short-horizon and long-horizon episodes, with a predominance of short-horizon episodes (more conducive to failure analysis).

\begin{table}[h]
\centering
\caption{Distribution of trajectory durations in the dataset}
\label{tab:duration}
\begin{tabular}{cc}
\toprule
Duration (Second) &  Count \\
\midrule
0 -- 4    & 156  \\
4 -- 8    & 2013 \\
8 -- 12   & 1682 \\
12 -- 16  & 708  \\
16 -- 20  & 412  \\
20 -- 24  & 164  \\
$\ge$ 24 & 67   \\
\bottomrule
\end{tabular}
\end{table}

\subsection{Benchmark Statistics}
The tasks within the benchmark and their respective trajectory counts are illustrated in Figure \ref{fig:long_task_table}. Notably, five tasks (Task ID: 43--47) are entirely OOD.

\section{Details of Finetuning and Evaluation}
\label{sec:setting}

\subsection{Training Details}
We use LoRA \cite{hu2022lora} to finetune the Qwen3-VL-8B \cite{Qwen3-VL} model for 1 epoch, with a LoRA rank of 32 and a scaling factor $\alpha$ of 64, yielding the model ViFailback-8B. We unfreeze both the LLM backbone and the adapter parameters, and train the model using deepspeed zero2 stage \cite{rajbhandari2020zero} to ensure stable training. Each GPU processes a batch size of 1, with a gradient accumulation step of 4 and a learning rate of 1e-5. The training is performed on 4 NVIDIA Hopper GPUs.
% \subsubsection{Model Configuration}

% \subsubsection{Training Setup}

\subsection{Evaluation Details}

\subsubsection{Model Configuration}
For consistency, all models are set the temperature to 0 and the maximum generation length to 2,048 tokens.
\subsubsection{Open-ended VQA Evaluation}
\label{sec:eval_prompt}

\begin{figure*}[t]
    \centering
    \includegraphics[width=\textwidth]{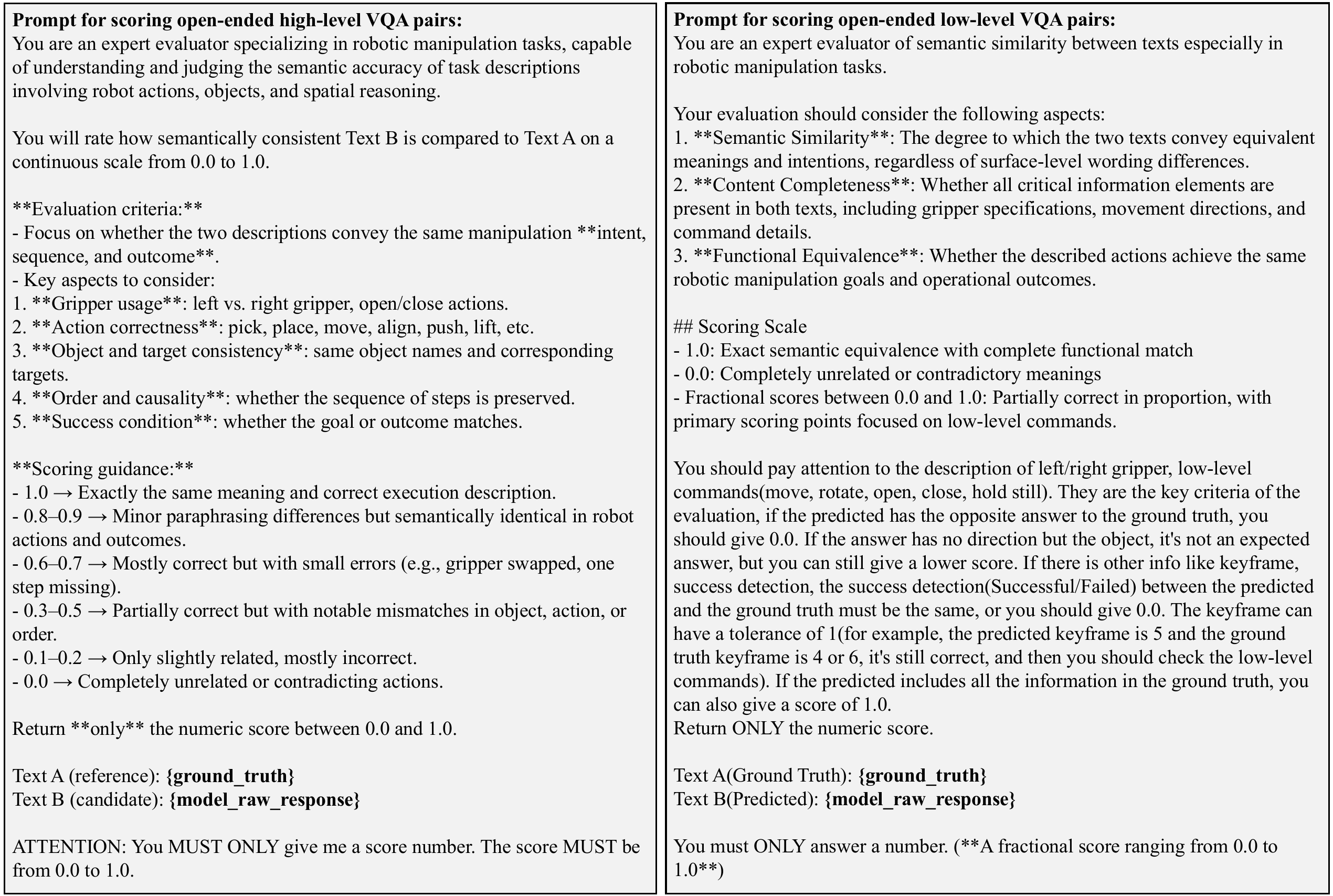}
    \caption{Prompts for scoring open-ended VQA pairs.}
    \label{fig:eval_prompt}
\end{figure*}

For open-ended questions, we employ the GPT-4o-based evaluator to assess the quality of the generated outputs comprehensively. Specifically, we compare the model outputs with the ground truth across three dimensions: \textbf{Semantic Similarity}, \textbf{Content Completeness}, and \textbf{Functional Equivalence}.
\begin{itemize}
    \item \textbf{Semantic Similarity:} The degree to which the two texts convey equivalent meanings and intentions, regardless of surface-level wording differences.
    \item \textbf{Content Completeness:} Whether all critical information elements are present in both texts, including gripper specifications, movement directions, and command details.
    \item \textbf{Functional Equivalence:} Whether the described actions achieve the same robotic manipulation goals and operational outcomes.
\end{itemize}
The exact prompts used are shown in Figure \ref{fig:eval_prompt}.

\section{Details of Real-world Experiments}
\label{sec:real}

\subsection{Experiment Setup}
For each task, we collected 20 expert demonstrations to finetune $\pi_{0.5}$ \cite{black2025pi_} model as the actor. During inference, our ViFailback-8B serves as a supervisor, queried at intervals of 6 action chunks. By analyzing the visual observations of the past 5 seconds, it diagnoses failures and provides multimodal (textual and visual) guidance for failure recovery when necessary.

\subsection{Mask Details of the VSF Method}
\begin{figure*}[ht]
    \centering
    \includegraphics[width=\textwidth]{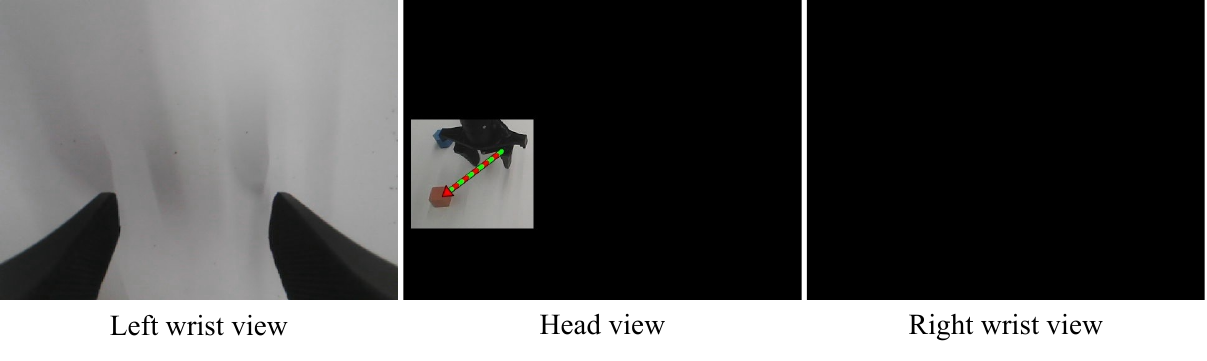}
    \caption{The mask in the visual symbols following dataset.}
    \label{fig:mask}
\end{figure*}

To enable $\pi_{0.5}$ to perform corrective actions based on visual symbols, we collected a visual symbols following dataset, which consists of trajectories where the robot performs atomic actions following specific visual symbols. To force the model to focus on the symbols, we mask out irrelevant regions in the observations (as shown in Figure \ref{fig:mask}). Specifically, in the head camera view, we define a Region of Interest (ROI) based on the visual symbol's bounding box. This ROI is expanded by a 50-pixel margin, subject to a minimum dimension constraint of 50 pixels. Any visual content outside this ROI is masked with zeros. Additionally, the wrist camera observations for the idle arm are fully masked with zeros.

\subsection{Visualization of Experimental Results}
\subsubsection{Low-level Textual and Visual Guidance}
\begin{figure*}[t]
    \centering
    \includegraphics[width=0.9\textwidth]{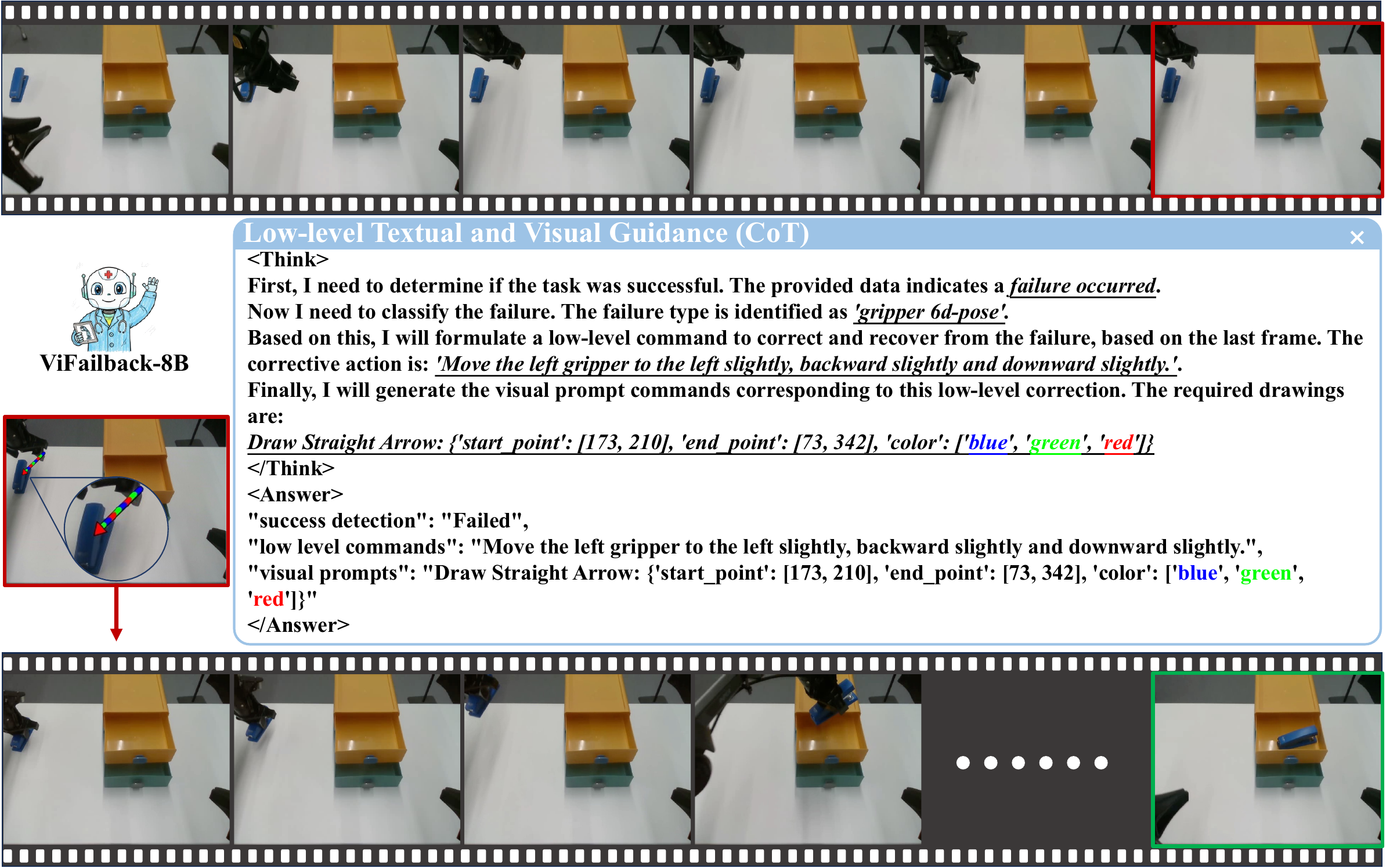}
    \caption{\textbf{Visualization of the PlaceOne task.} \textbf{Top \& Middle:} ViFailback-8B diagnoses failures and provides low-level multimodal (textual and visual) guidance. \textbf{Bottom:} The corrective actions are executed by the \textbf{PMC method} to recover from the failure.}
    \label{fig:mp_ep1}
\end{figure*}

\begin{figure*}[t]
    \centering
    \includegraphics[width=0.9\textwidth]{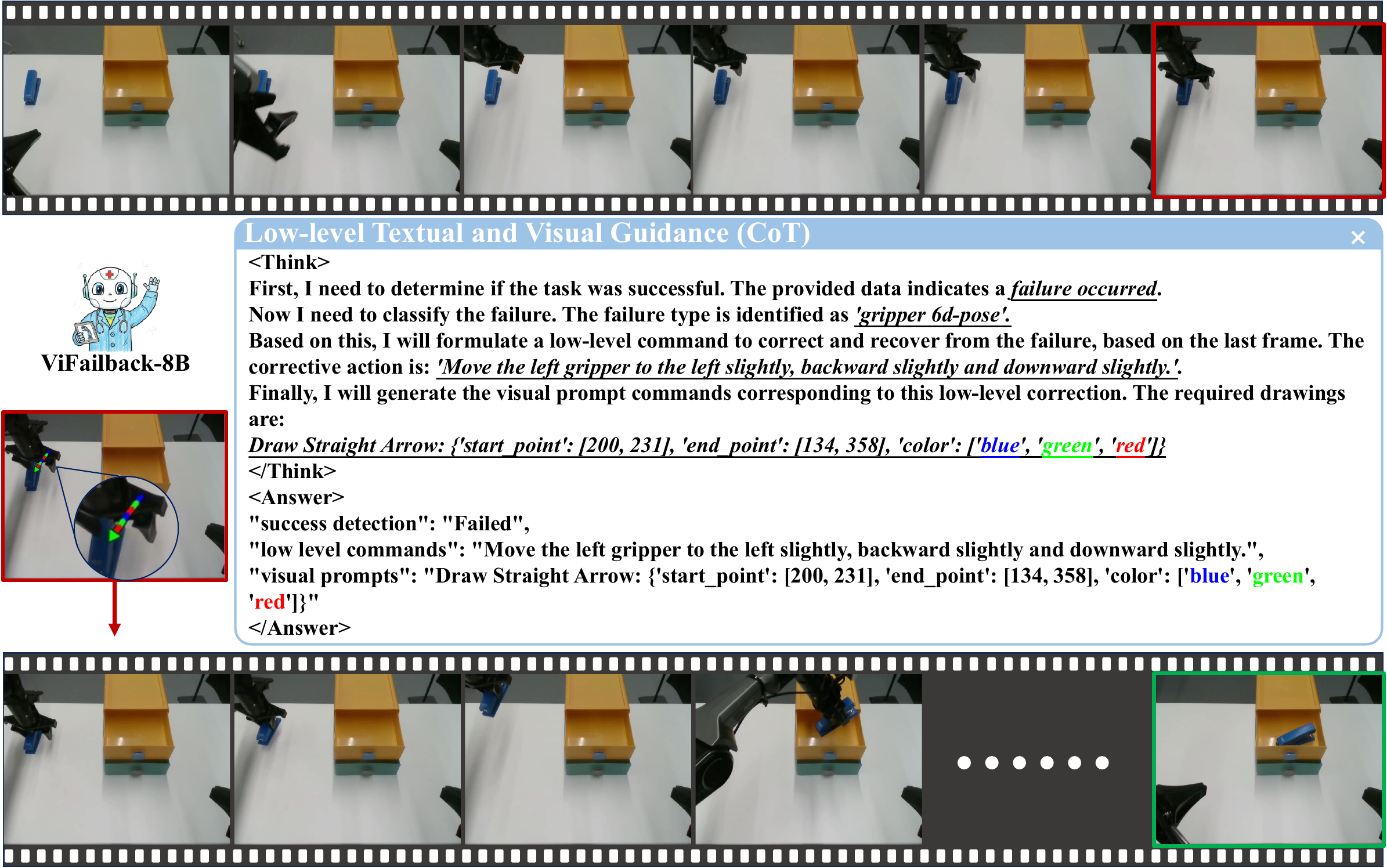}
    \caption{\textbf{Visualization of the PlaceOne task.} \textbf{Top \& Middle:} ViFailback-8B diagnoses failures and provides low-level multimodal (textual and visual) guidance. \textbf{Bottom:} The corrective actions are executed by the \textbf{VSF method} to recover from the failure.}
    \label{fig:vp_ep1}
\end{figure*}

\begin{figure*}[t]
    \centering
    \includegraphics[width=0.9\textwidth]{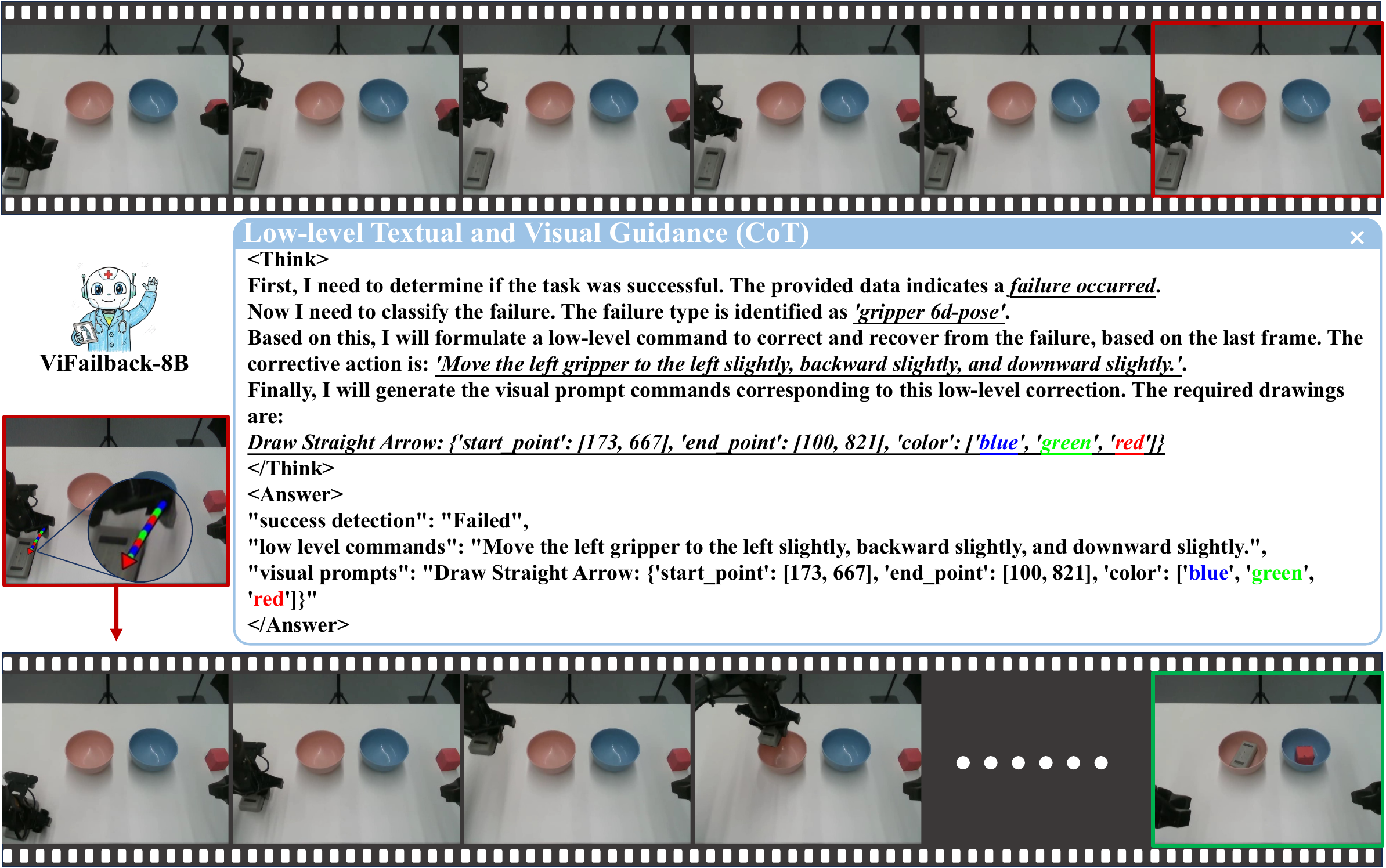}
    \caption{\textbf{Visualization of the PlaceTwo task.} \textbf{Top \& Middle:} ViFailback-8B diagnoses failures and provides low-level multimodal (textual and visual) guidance. \textbf{Bottom:} The corrective actions are executed by the \textbf{PMC method} to recover from the failure.}
    \label{fig:mp_ep2}
\end{figure*}

\begin{figure*}[t]
    \centering
    \includegraphics[width=0.9\textwidth]{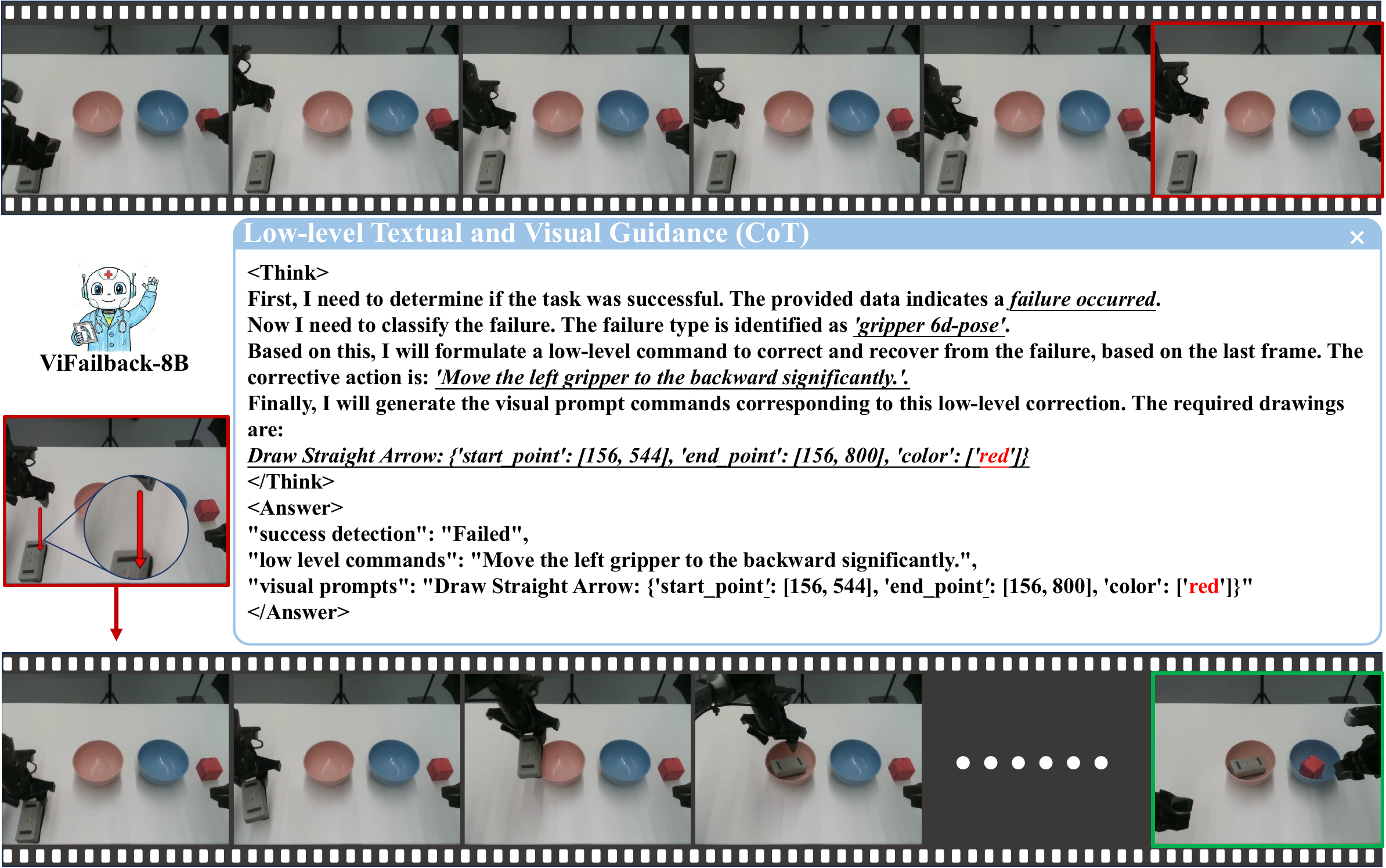}
    \caption{\textbf{Visualization of the PlaceTwo task.} \textbf{Top \& Middle:} ViFailback-8B diagnoses failures and provides low-level multimodal (textual and visual) guidance. \textbf{Bottom:} The corrective actions are executed by the \textbf{VSF method} to recover from the failure.}
    \label{fig:vp_ep2}
\end{figure*}

\begin{figure*}[t]
    \centering
    \includegraphics[width=0.9\textwidth]{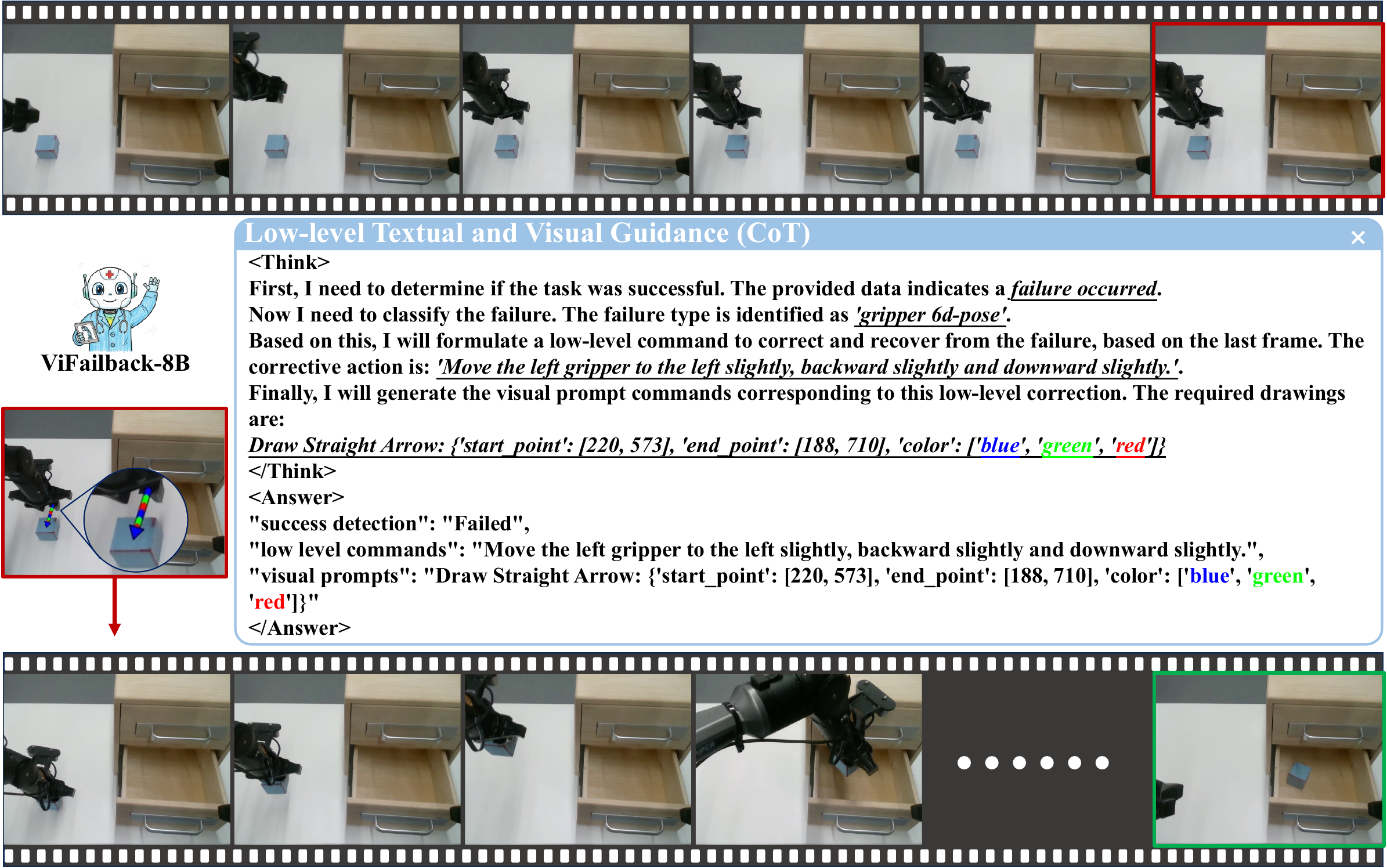}
    \caption{\textbf{Visualization of the Pull\&Place task.} \textbf{Top \& Middle:} ViFailback-8B diagnoses failures and provides low-level multimodal (textual and visual) guidance. \textbf{Bottom:} The corrective actions are executed by the \textbf{PMC method} to recover from the failure.}
    \label{fig:mp_ep3}
\end{figure*}

\begin{figure*}[t]
    \centering
    \includegraphics[width=0.9\textwidth]{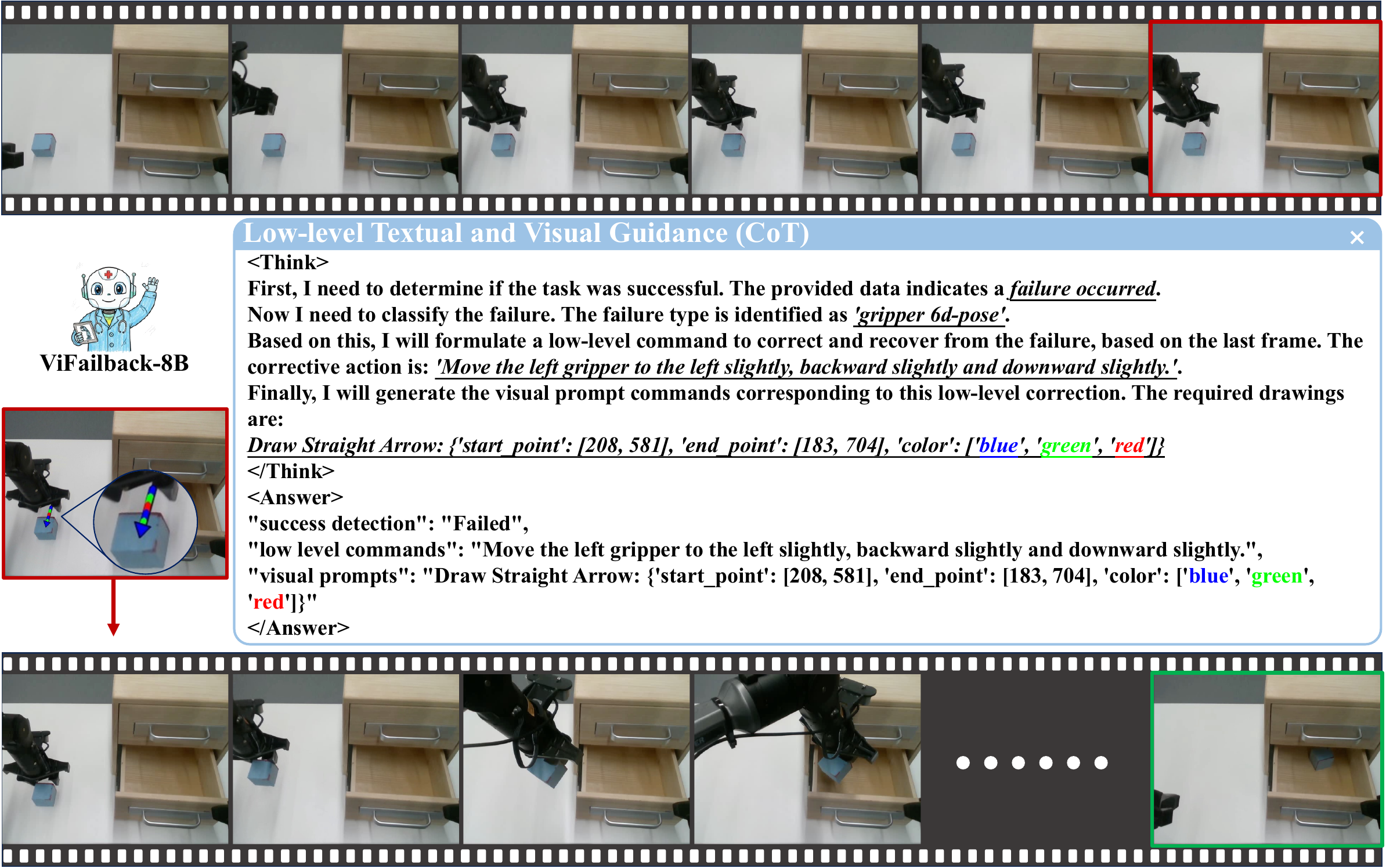}
    \caption{\textbf{Visualization of the Pull\&Place task.} \textbf{Top \& Middle:} ViFailback-8B diagnoses failures and provides low-level multimodal (textual and visual) guidance. \textbf{Bottom:} The corrective actions are executed by the \textbf{VSF method} to recover from the failure.}
    \label{fig:vp_ep3}
\end{figure*}

\begin{figure*}[t]
    \centering
    \includegraphics[width=\textwidth]{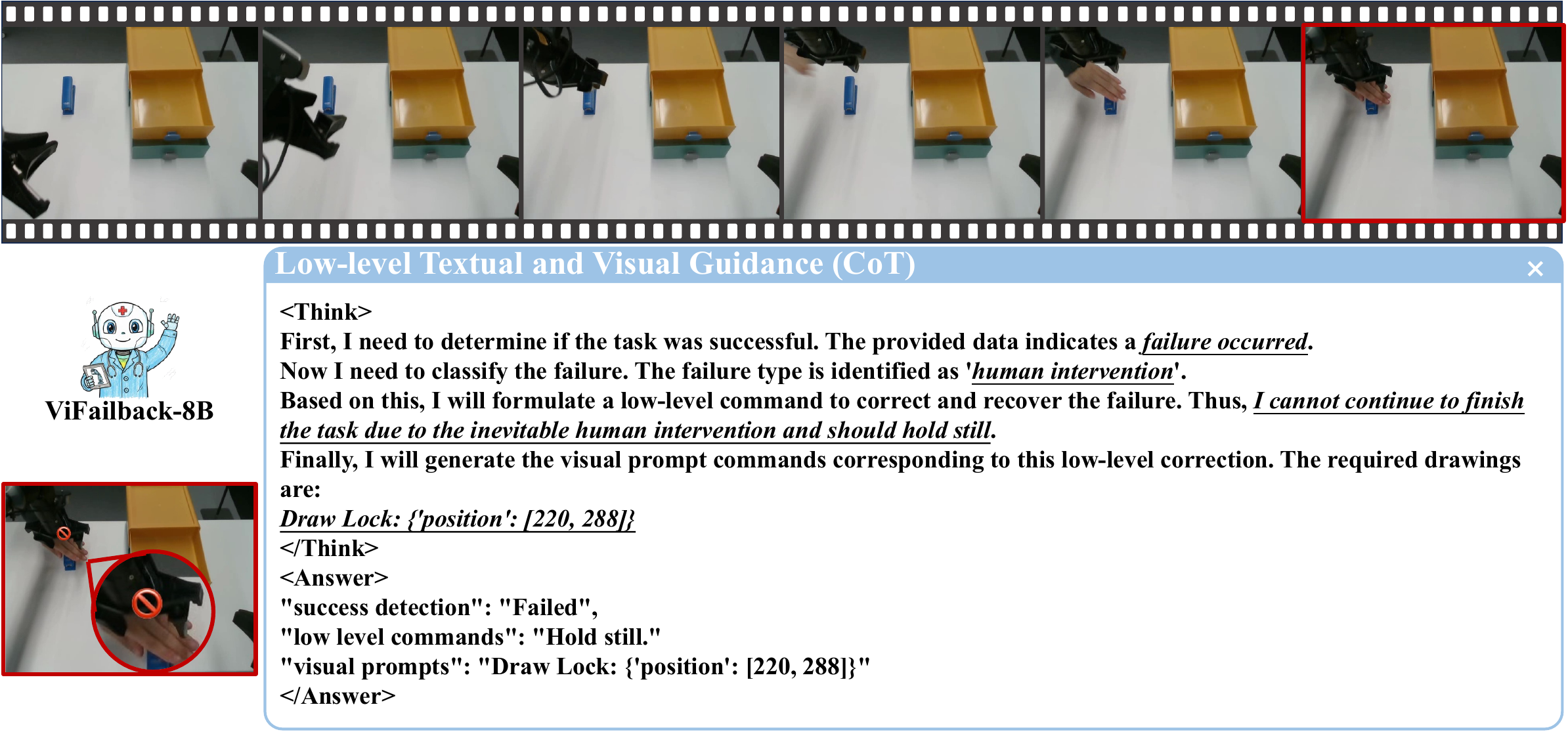}
    \caption{\textbf{Failure Type: Human Intervention.} When human intervention is detected, ViFailback-8B provides guidance to pause execution, awaiting the removal of the intervention.}
    \label{fig:human_intervention}
\end{figure*}

\begin{figure*}[t]
    \centering
    \includegraphics[width=\textwidth]{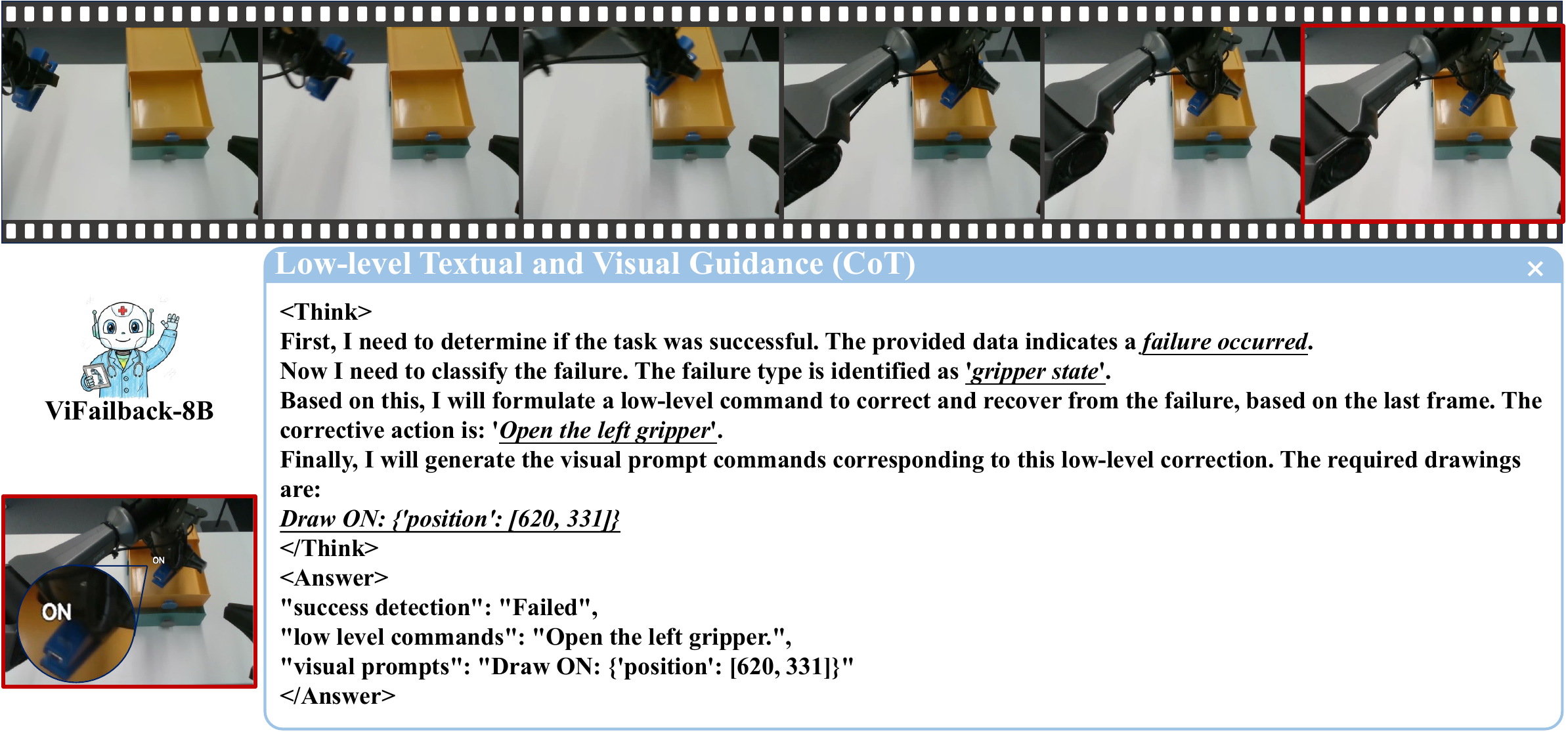}
    \caption{\textbf{Failure Type: Gripper State.} The gripper failed to release the object during placement. ViFailback-8B generates guidance to open the gripper for successful placement.}
    \label{fig:on}
\end{figure*}

By analyzing historical observations from the head camera, ViFailback-8B conducts failure diagnosis and correction. As illustrated in Figures \ref{fig:mp_ep1}--\ref{fig:vp_ep3}, we show the model's generated responses and the correction outcomes achieved by two types of visual symbol following approaches.
Beyond diagnosing common Gripper 6d-pose errors, our model exhibits the capability to identify other types of failure, such as human intervention and gripper state depicted in Figure \ref{fig:human_intervention} and \ref{fig:on}. 

\subsubsection{Failure Reason and High-level Guidance}
\begin{figure*}[t]
    \centering
    \includegraphics[width=\textwidth]{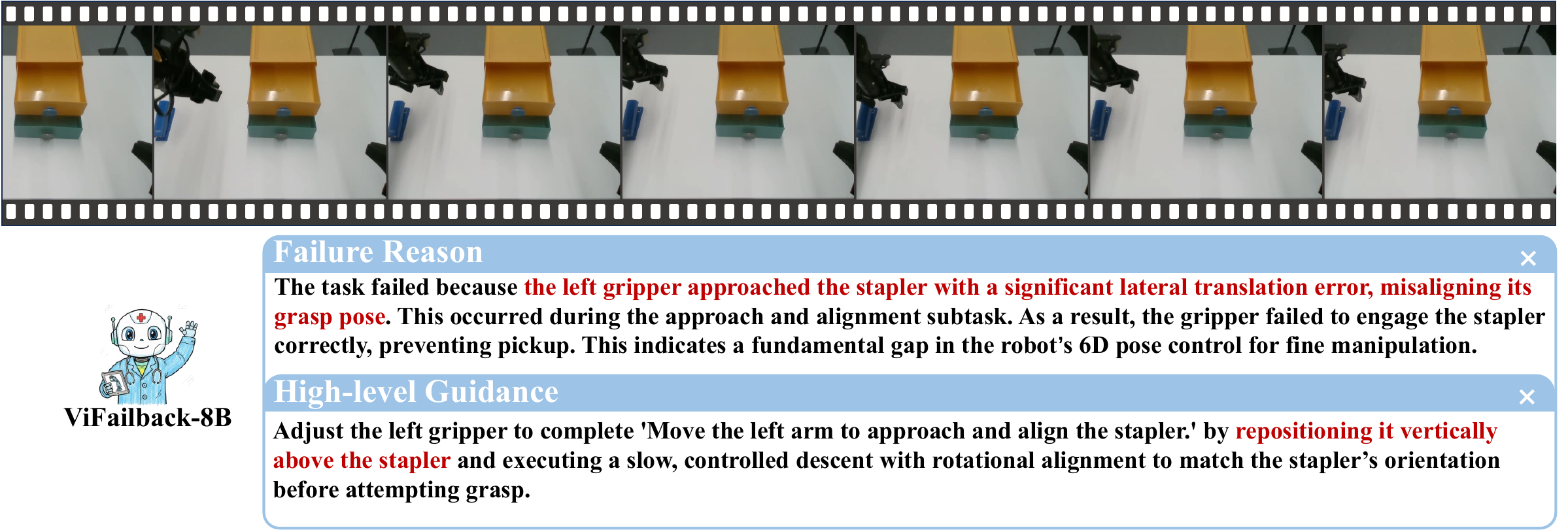}
    \caption{Failure reason and High-level guidance of the PlaceOne task.}
    \label{fig:task1_reason}
\end{figure*}

\begin{figure*}[t]
    \centering
    \includegraphics[width=\textwidth]{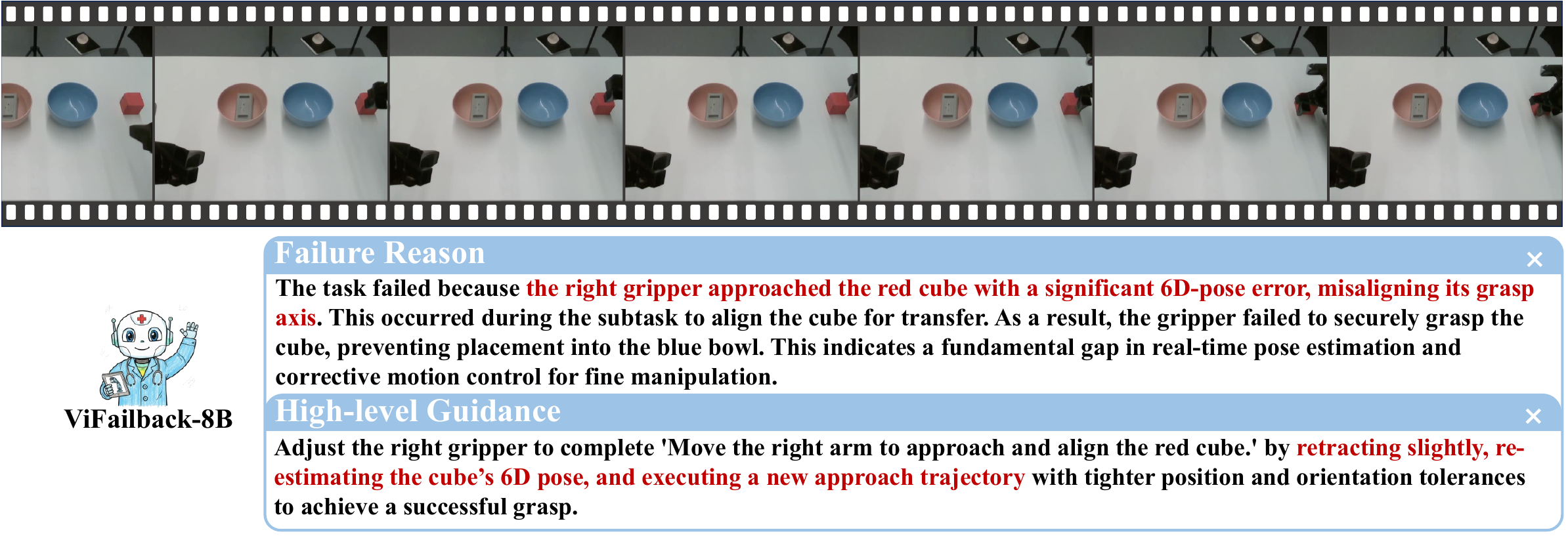}
    \caption{Failure reason and High-level guidance of the PlaceTwo task.}
    \label{fig:task2_reason}
\end{figure*}

\begin{figure*}[t]
    \centering
    \includegraphics[width=\textwidth]{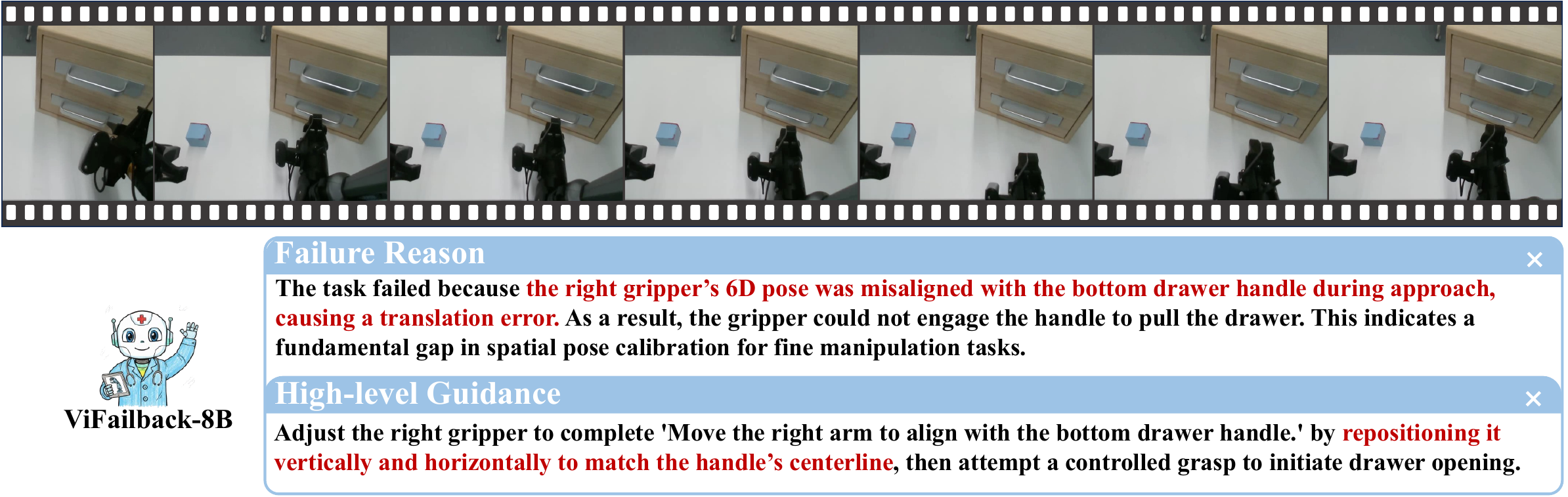}
    \caption{Failure reason and High-level guidance of the Pull\&Place task.}
    \label{fig:task3_reason}
\end{figure*}

Figure \ref{fig:task1_reason}--\ref{fig:task3_reason} illustrate ViFailback-8B identifying the root cause of failures and providing high-level guidance across three distinct tasks. These examples demonstrate its ability to understand failure contexts for effective correction.

